\newif\ifarxiv

\arxivtrue

\ifarxiv
\documentclass[10pt]{article}
\usepackage[margin=1in]{geometry}
\else
\documentclass[journal]{IEEEtran}
\fi

\usepackage{color}

\def\fig#1{Fig.~\ref{fig:#1}}

\def\eq#1{(\ref{eq:#1})}
\def\tab#1{Table~\ref{tab:#1}}

\usepackage{cite}
\usepackage[pdftex]{graphicx} 

\usepackage{amsmath, amssymb, bm} 
\begin{document}
%

\title{A Supervised Learning Algorithm for Multilayer Spiking Neural Networks Based on Temporal Coding Toward Energy-Efficient VLSI Processor Design}
%
%
%

\author{Yusuke~Sakemi, 
        Kai~Morino, 
			Takashi~Morie, 
        and~Kazuyuki~Aihara,
\thanks{Y. Sakemi and K. Morino, and K. Aihara are with Institute of Industrial Science, The University of Tokyo, Tokyo 153-8505, Japan e-mail: sakemi@sat.t.u-tokyo.ac.jp.}
\thanks{Y. Sakemi is with NEC Corporation, Kanagawa 211-8666, Japan.}
\thanks{K. Morino is with Interdisciplinary Graduate School of Engineering Sciences, Kyushu University, 816-8580, Japan}
\thanks{T. Morie is with Graduate School of Life Science and Systems Engineering, Kyushu Institute of Technology, Kitakyushu, 808-0196, Japan.}
\thanks{K. Aihara is with International Research Center for Neurointelligence (WPI-IRCN), The University of Tokyo Institutes for Advanced Study, The University of Tokyo, Tokyo 113-0033, Japan.}
}

\ifarxiv
\renewcommand\footnotemark{}
\fi

%
%

\markboth{IEEE TRANSACTIONS ON NEURAL NETWORKS AND LEARNING SYSTEMS}%
{Shell \MakeLowercase{\textit{et al.}}: Bare Demo of IEEEtran.cls for IEEE Journals}
%



\maketitle

\begin{abstract}
Spiking neural networks (SNNs) are brain-inspired mathematical models with the ability to process information in the form of spikes.
SNNs are expected to provide not only new machine-learning algorithms, but also energy-efficient computational models when implemented in VLSI circuits.
In this paper, we propose a novel supervised learning algorithm for SNNs based on temporal coding.
A spiking neuron in this algorithm is designed to facilitate analog VLSI implementations with analog resistive memory, 
by which ultra-high energy efficiency can be achieved.
We also propose several techniques to improve the performance on a recognition task,
and show that the classification accuracy of the proposed algorithm is as high as that of the state-of-the-art temporal coding SNN algorithms on the MNIST dataset.
Finally, we discuss the robustness of the proposed SNNs against variations that arise from the device manufacturing process and are unavoidable in analog VLSI implementation.
We also propose a technique to suppress the effects of variations in the manufacturing process on the recognition performance.
\end{abstract}

\ifarxiv
\else
\fi

%
\ifarxiv
\else
\IEEEpeerreviewmaketitle
\fi

\section{Introduction}
%
%
%
%
\ifarxiv
Machine 
\else
\IEEEPARstart{M}{achine}
\fi
learning is a paradigm according to which algorithms perform tasks by learning from data instead of exploiting domain knowledge.
Recently, machine-learning algorithms that employ artificial neural networks (ANNs) have been attracting much interest from researchers and engineers in diverse fields owing to their astonishing performance on various tasks such as image recognition, natural language processing, and time series forecasting \cite{Krizhevsky2012imagenet, Lecun2015deep}.
These machine-learning algorithms are known as {\it deep learning} because of the characteristic multilayer structures of ANNs.
In deep learning, {\it stochastic gradient descent} (SGD) and {\it back propagation} (BP) are important techniques that enable multilayer ANNs to perform effective learning \cite{Amari1967theory,Rumelhart1988learning}. 
In addition, recent advances in algorithms such as {\it convolutional neural networks} (CNNs) \cite{Lecun1989backpropagation} and skip connections\cite{He2016deep} have led to the success of deep learning. 
Moreover, the evolution of very large-scale integration (VLSI) technologies, such as Moore's law \cite{Moore1965cramming} and Dennard scaling\cite{Dennard1974design}, have enabled deep learning to be implemented in real-world applications.

However, the computational burden of deep learning often hinders its application in the real world.
Therefore, researchers and engineers often use high-performance processors such as graphical processing units, tensor processing units\cite{Jouppi2017datacenter}, and application-specific integrated circuits (ASICs)\cite{Sze2017} to compensate for this shortcoming.
Despite these developments, it remains difficult to enable the learning to converge in an acceptable time, or to perform the tasks in real time.
This is even more difficult when performing tasks with processors placed near the end-users (edge computing) owing to the limited battery capacity.
Therefore, for edge computing, high computational energy efficiency is required.

In this regard, finding a solution for the energy-efficiency problem by looking toward the human brain would only seem natural, because the human brain is capable of processing complex information in real time by consuming only 20-30 W of energy.
An ANN is mostly a mathematical model inspired by the physiological observation of spike frequencies in biological neurons,
whereas a spiking neural network (SNN) is a mathematical model that is directly concerned with the spike generation \cite{Maass2015tospike}.
Therefore, SNNs would be expected to perform tasks in a way similar to the brain with low energy consumption.
Until recently, SNNs did not perform as well as ANNs on tasks such as image recognition. 
However, it has been shown that the performance of SNNs could reach that of ANNs by exploiting the techniques used in deep learning (multilayer structures, SGD, BP, etc.)\cite{Tavanaei2019Deep, Pfeiffer2018Deep}. 
Thus, it seems reasonable to find a way to really improve the energy efficiency of SNNs.
In fact, not only have SNNs demonstrated their high algorithmic performance but also their advantages in terms of energy efficiency especially when implemented in ASICs by exploiting the asynchronous property and binary (spike) communications \cite{Liu2014event}.
Many research groups have manufactured various types of ASICs for SNNs, which include, for example, mixed-signal ASICs\cite{Qiao2015reconfigurable, Moradi2018scalable, Schemmel2010, Friedmann2017demonstrating}, and fully digital ASICs \cite{Merolla2014, Davies2018loihi}.
Other ASICs were also reported \cite{Schuman2017survey, Bouvier2019spiking}.
These ASICs are able to execute SNNs with complex and flexible networks in a highly energy efficient manner.
However, it is still unclear whether ASICs for SNNs are more energy efficient per operation or per task than ASICs for ANNs.
This comparison is blurred by a common bottleneck preventing computational energy efficiency, which is caused by data movement between processors and memory \cite{Horowitz2014computing}. 
Because of this common bottleneck, the advantages of employing SNNs in a hardware implementation such as the asynchronous property and spike communications are hardly visible in terms of the computational energy efficiency of ASICs.
One promising approach to overcome this bottleneck is {\it in-memory computing}.

In-memory computing is a concept that involves performing the computing where the memorized data are located \cite{Verma2019inmemory}.
Especially, in-memory computing schemes that use analog resistive memory have being attracting hardware researchers' attention \cite{Tsai2018recent, Ielmini2018memory, Xia2019memristive}.
Analog resistive memory stores information in the form of resistance.
The application of voltage across the two terminals of an analog resistive memory cell results in an electric current, which is proportional to the voltage, and which flows through the memory cell by obeying Ohm's law.
Currents flowing through memory cells can easily be summed up on the basis of Kirchhoff's current law.
Note that this computing scheme does not require information to be stored elsewhere (e.g., static random access memory or dynamic random access memory).
Based on the above principle, computing such as matrix-vector multiplications (MVMs) can be performed in a highly energy efficient way.

This motivated us to propose a novel supervised learning algorithm for multilayer SNNs to achieve high algorithmic performance and high energy efficiency. 
It is also possible to implement our algorithm in VLSI circuits with analog resistive memory.
Our algorithm is based on the following ideas: (i) SNNs are composed of spiking neuron models which facilitate VLSI implementations;
(ii) SNNs are trained with a temporal coding scheme, thereby enabling SNNs to perform tasks with fewer spikes than SNNs based on rate coding.
We also propose novel learning techniques to improve the learning performance  
and evaluate the performance on the MNIST dataset.
Furthermore, for real VLSI implementations, robustness against manufacturing variations in devices is crucial.
Therefore, we investigated the effects of these manufacturing variations on the learning performance of the proposed algorithm.
In addition, we propose a technique to suppress the effect of manufacturing variations.

This paper consists of the following sections:
section \ref{ss:related_works} describes related work;
section \ref{ss:model_learning} proposes an SNN and its learning algorithm;
section \ref{ss:experimental_setup} introduces the experimental setup for evaluating the algorithm on the MNIST dataset, and for investigating the effects of manufacturing variations;
section \ref{ss:results} presents the experimental results;
section \ref{ss:conclusion} concludes this paper.

\section{Related Work} \label{ss:related_works}
\subsection{Supervised learning algorithms for multilayer SNNs}

\begin{figure}[!t]
\centering
\includegraphics[width=8cm]{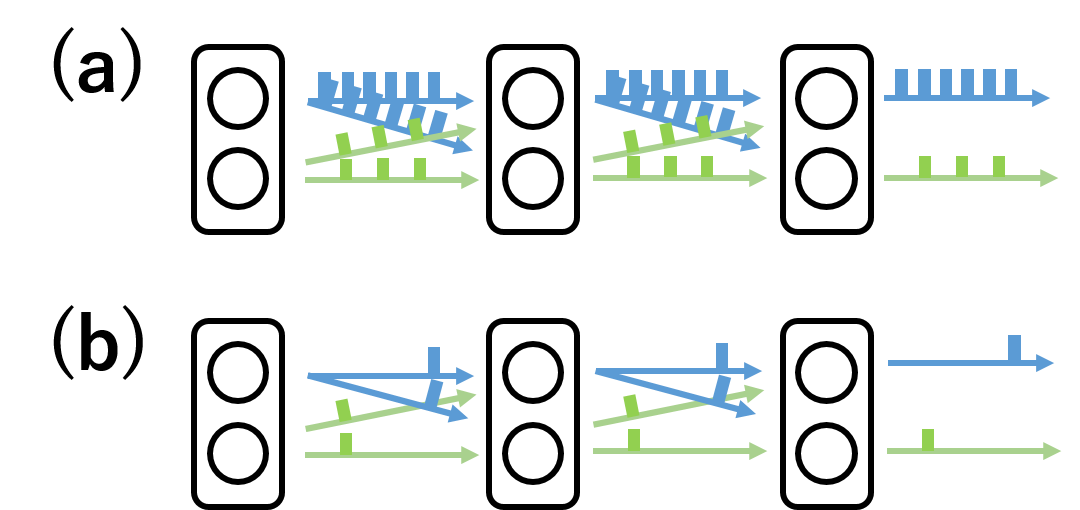} 
\caption{
Different representations of information in SNNs. The arrows represent the directions of information propagation. The widths of spikes do not carry information.
(a) Information is contained in the spike frequency (rate coding). (b) Information is contained in the spike timing (temporal coding). 
}
\label{fig:comparison}
\end{figure}

A number of studies have shown that multilayer SNNs can be trained with the BP technique \cite{Tavanaei2019Deep, Pfeiffer2018Deep, Neftci2019surrogate}.
The application of the BP to SNNs is not straightforward as in the case of ANNs because SNNs have non-differentiable processes such as spike generations. 
Furthermore, the dynamics and the learning performance depend on the coding schemes of SNNs. 
\fig{comparison} shows SNNs based on two different coding schemes: rate coding and temporal coding.
In rate coding, information is contained in the spike frequency of neurons, whereas in temporal coding, information is contained in the spike timing of neurons \cite{Maass1999}.

In rate coding, BP algorithms can be obtained by approximations such as linear neuron approximation \cite{Ponulak2010supervised}, 
and by considering spike generation as a probabilistic process \cite{Neftci2018event}. 
In \cite{Lee2016training, Zenke2018superspike, Neftci2019surrogate}, the BP algorithms were obtained by using the derivative of the membrane potential with respect to time. 
In addition, it has been shown that trained ANNs can be converted into rate-coding-based SNNs \cite{Diehl2015fast,Esser2015energy}.
However, these algorithms need to average a number of spikes to propagate information forward or backward. 
Because the energy consumption of ASICs increases as the number of spikes increases, performing tasks with fewer spikes is preferable.

Temporal coding enables SNNs to perform tasks with fewer spikes compared to rate coding.
SpikeProp is the first algorithm that enabled multilayer SNNs to learn by using temporal coding\cite{Bohte2002error}.
SpikeProp obtains the BP algorithm by directly differentiating spike timing with respect to weights.
Mostafa \cite{Mostafa2018supervised} improved the learning performance of SNNs in temporal coding on recognition tasks by using non-leaky spiking neurons. 
Comsa \textit{et al.} \cite{Comsa2019temporal} used an alpha synaptic function and employed an evolutionary algorithm to search for the hyperparameter to improve the learning performance.

In this paper, we propose a supervised learning algorithm for SNNs based on temporal coding, which is similar to those mentioned above \cite{Mostafa2018supervised, Comsa2019temporal}.
The novelties of our work are as follows: we use a different neuron model to simplify the hardware implementation and modify the learning algorithm to improve the learning performance.

\subsection{Computing with analog resistive memory}
 
Analog resistive memory such as resistive RAM and phase change memory is considered important building blocks for next-generation artificial-intelligence processors \cite{Tsai2018recent, Ielmini2018memory, Xia2019memristive}.
Many research groups have already designed circuits with analog resistive memory to efficiently compute MVMs \cite{Burr2015experimental, Choi2017experimental, Bayat2018implementation, Li2018analogue, Cai2019fully}. 
Among various VLSI implementations with analog resistive memory, circuits that process information by using the timing of electronic pulses are expected to be advantageous in terms of energy efficiency because of their simple implementation \cite{Cai2019fully, Ravinuthula2009time,Shafiee2016isaac, Marinella2018multi, Ambrogio2018equivalent, Wang2016, Wang2018time, Bavandpour2019energy, Yamaguchi2019energy}.
For example, circuits using the timing difference between two pulses \cite{Wang2016,Wang2018time}  
and those using the pulse-width difference between two pulses \cite{Bavandpour2019energy, Yamaguchi2019energy}  
are able to compute MVMs efficiently.
Nandakumar \textit{et al.} \cite{Nandakumar2019supervised} developed circuits with phase change memory that is capable of computing single-layer SNNs based on rate coding.

In this paper, we show that SNNs based on temporal coding can be straightforwardly mapped to the above VLSI implementation schemes.

\section{Methods}\label{ss:model_learning}
This section introduces our proposed SNNs and their learning algorithms.
The SNNs described here are feedforward networks of spiking neurons as shown in \fig{comparison}(b), where spikes propagate from one layer to the next layer without skipping layers.
The SNNs are to be trained such that the network can convert a given input spike pattern into a specific output spike pattern.
We also present several examples of VLSI implementations of the SNNs where analog resistive memory is used as network weights.

\subsection{Models}

\begin{figure}
	\begin{center}
		\includegraphics[clip, width=8cm]{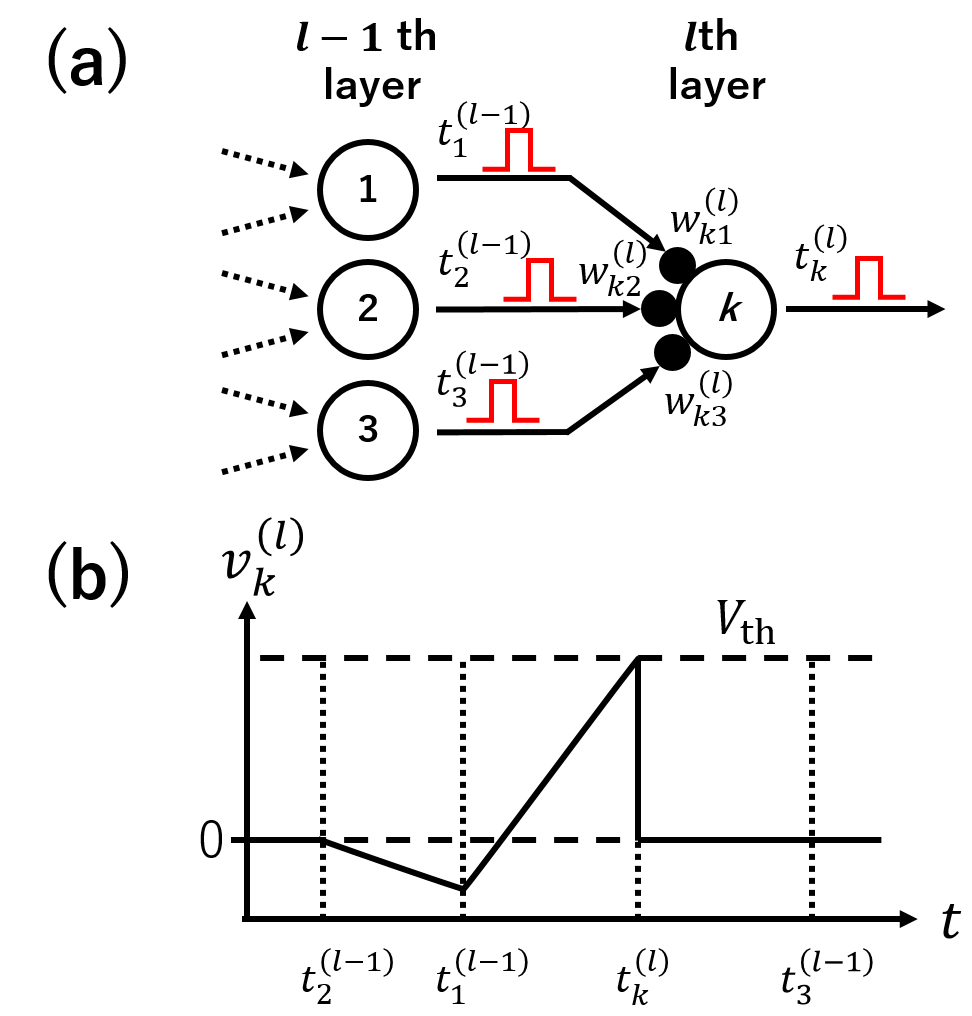}
		\caption{(a) Schematic diagram of a multilayer SNN. 
The $k$th neuron in the $l$th layer receives spikes from neurons in the $(l-1)$th layer at $t_1^{(l-1)}, t_2^{(l-1)}, t_3^{(l-1)}$, 
and fires at $t_k^{(l)}$.
Then, a spike is sent to neurons in the subsequent layer. 
(b) Schematic diagram of time evolution of the membrane potential $v_k^{(l)}$ of the $k$th neuron in the $l$th layer.
}
		\label{fig:IF_model}
	\end{center}
\end{figure}

The time evolution of the membrane potential of a spiking neuron that we employed is given by
\begin{flalign}
\frac{d }{dt} v_i^{(l)}(t) &= \sum _{j=1} ^{N^{(l-1)}} w_{ij}^{(l)} \theta(t - t_j^{(l-1)}) \label{eq:neuron} \\
\theta(x) &= \begin{cases}
0, & ~(x<0) \\
1, & ~(x\ge 0)
\end{cases} 
\end{flalign}
where we define $v_i^{(l)}(t)$ as the membrane potential of the $i$th spiking neuron in the $l$th layer,
and define $t^{(l)}_i$ as the spike timing generated by the same neuron to the present input pattern.
We define $w_{ij}^{(l)}$ as the weight of the connection from the $j$th neuron in the $(l-1)$th layer to the $i$th neuron in the $l$th layer.
$N^{(l)}$ is the number of neurons constituting the $l$th layer.
The zeroth layer and the $M$th layer are referred to as the input and output layers, respectively.
The other layers are hidden layers.
This network is an $M$-layer SNN.

The analytical solution of \eq{neuron} is given by 
\begin{flalign}
v_i^{(l)}(t) = \sum _{j=1} ^{N^{(l-1)}} w_{ij}^{(l)}(t - t_j ^{(l-1)})\theta(t - t_j ^{(l-1)}).
\end{flalign}
The neuron generates (fires) a spike when its membrane potential reaches the firing threshold $V_\text{th}$.
Then, the membrane potential is fixed to the reset voltage ($v_i^{(l)}=0$), which prevents the neuron from firing twice for a given input spike pattern. 
\fig{IF_model}(b) shows a schematic diagram to indicate the dependence of the evolution of the membrane potential on the timing of received spikes and the connection weights. 

The timing of spikes can be obtained as follows.
The spike timing $t_i ^{(l)}$ satisfies the equation
\begin{align}
V_\text{th} = \sum _{j\in\Gamma _i^{(l)}} w_{ij}^{(l)} (t_i^{(l)}-t_j^{(l-1)}) \label{eq:firing_condition} 
\end{align}
where $\Gamma _i^{(l)}$ is a set of indices of neurons in the $(l-1)$th layer that sends spikes before the $i$th neuron in the $l$th layer fires.
With \eq{firing_condition}, we obtain the spike timing as follows:
\begin{align}
t_i ^{(l)} &= \frac{V_\text{th} + \sum _{j\in \Gamma _i^{(l)}} w_{ij}^{(l)} t_j^{(l-1)} }{\sum _{j\in \Gamma _i^{(l)}} w_{ij}^{(l)}}. \label{eq:SpikeTime}  
\end{align}

\subsection{Supervised learning of SNNs}

For the supervised learning of SNNs, we define the following functions
\begin{flalign}
C(\bm{t}^{(M)}, \bm{\kappa};\bm{w}) &:= L(\bm{t}^{(M)}, \bm{\kappa};\bm{w})  + \frac{\gamma}{2}  R(\bm{t}^{(M)}) \label{eq:Cost} \\
L(\bm{t}^{(M)}, \bm{\kappa};\bm{w}) &:=  -\sum _{i=1} ^{N^{(M)}} \kappa _i\ln \left(S_i (\bm{t}^{(M)})\right) \label{eq:CCE_Softmax} \\
S_i (\bm{t}^{(M)})&:= \frac{\exp \left( -t_i^{(M)} \right)}{\sum _{j=1} ^{N^{(M)}} \exp \left( -t_j^{(M)} \right)} \label{eq:softmax}\\
R(\bm{t}^{(M)}) &:=   \sum_{i=1} ^{N^{(M)}}\left( t_i ^{(M)} - t^\text{(Ref)} \right)^2 \label{eq:temporal_penalty}
\end{flalign}
where $C(\cdot, \cdot; \cdot),~L(\cdot,\cdot;\cdot),~S_i(\cdot), \text{ and }R(\cdot)$ are the cost function, loss function, softmax function, and temporal penalty term, respectively.
Teacher labels are represented as a vector $\bm{\kappa}$ such that $\kappa_i$ is 1 when the $i$th label is given, and 0, otherwise.
The weights of the entire network are represented as a vector $\bm{w}$.
The temporal penalty term $R(\bm{t}^{(M)})$ is defined by the total difference between the spike timing of the output neurons $\bm{t}^{(M)}$ and the timing of a reference spike $t^\text{(Ref)}$.
Note that $\bm{t}^{(M)}$ is the vector of spike timing of the neurons in the output layer.
As explained later, the temporal penalty is intended to circumvent learning difficulties.
The coefficient $\gamma (>0)$ controls the effect of the temporal penalty term.

The network is trained by updating the weights by using SGD on a dataset as follows:
\begin{flalign}
\Delta w_{jk}^{(l)} = - \eta \frac{\partial C(\bm{t}^{(M)}, \bm{\kappa};\bm{w})}{\partial w_{jk}^{(l)}} \label{eq:SGD}
\end{flalign}
where $\Delta w_{jk}^{(l)}$ is the update of $w_{jk}^{(l)}$ and $\eta (>0)$ is the learning rate.
The derivative of the cost function with respect to weights is given by
\begin{flalign}
\frac{\partial C}{\partial w_{ij} ^{(l)}} = \frac{\partial t_i^{(l)}}{\partial w_{ij}^{(l)}} \delta _i ^{(l)}
\end{flalign}
where we define the propagation error $\delta _i ^{(h)}$ as
\begin{flalign}
\delta _i ^{(h)} =\begin{cases}
\sum _{k=1} ^{N^{(h+1)}} \frac{\partial t_k ^{(h+1)}}{\partial t_i^{(h)}} \delta _k ^{(h+1)},  \text{ for }h=1,2,\dots,M-1 \\
\frac{\partial C}{\partial t_i^{(M)}}, \text{ for }h=M.
\end{cases}
\end{flalign}
By using the following derivatives
\begin{flalign}
\frac{\partial t_i^{(l)}}{\partial w_{ij}^{(l)}} &= -\frac{t_i ^{(l)} - t_j ^{(l-1)}}{\sum _{k\in\Gamma _i^{(l)}}  w_{ik}^{(l)}} \label{eq:Denom1} \\
\frac{\partial t_i ^{(l)}}{\partial t_j^{(l-1)}} &= \frac{w_{ij}^{(l)}}{\sum _{k\in\Gamma _i^{(l)}}  w_{ik}^{(l)}} \label{eq:Denom2} \\
\frac{\partial C}{\partial t_i^{(M)}} &=  \kappa_i - S_i +\gamma \left( t_i ^{(M)} - t^\text{(Ref)} \right) \label{eq:Denom3}
\end{flalign}
the updating difference \eq{SGD} can be obtained.

The derivative of the cost function is rigorous unless a neuron ceases to fire because of the update.  
However, we found that the network experience two difficulties during training.
These difficulties are related to destructively large weight updates and non-unique spike timing, respectively.
The destructively large weight updates are caused by infinitesimal values in the denominator of \eq{Denom1} or \eq{Denom2}.
Our solution to prevent these unacceptable large weight updates is to redefine \eq{Denom1} and \eq{Denom2} as
\begin{flalign}
\frac{\partial t_i^{(l)}}{\partial w_{ij}^{(l)}} &= - \frac{t_i ^{(l)} - t_j ^{(l-1)}}{\epsilon + \sum _{k\in\Gamma _i^{(l)}}  w_{ik}^{(l)}} \label{eq:suppression1} \\
\frac{\partial t_i ^{(l)}}{\partial t_j^{(l-1)}} &= \frac{w_{ij}^{(l)}}{\epsilon + \sum _{k\in\Gamma _i^{(l)}}  w_{ik}^{(l)}} \label{eq:suppression2} 
\end{flalign}
where $\epsilon$ is a positive constant.

Next, the loss function shown in \eq{CCE_Softmax} is invariant against an arbitrary value $\alpha$ such that 
\begin{flalign}
L(\bm{t}^{(M)}, \bm{\kappa};\bm{w}) = L(\bm{t}^{(M)} + \alpha \bm{I}, \bm{\kappa};\bm{w})
\end{flalign}
where $\bm{I}\in \mathbb{R}^{N^{(M)}}$ is a vector of which the elements are all unity.
Owing to its invariant nature, the loss function does not uniquely specify the timing of output spikes, 
which forces the timing of the output spike to become large, resulting in the disappearance of spikes.
The temporal penalty term, which is introduced to solve this problem, 
enables the cost function to uniquely specify the timing of the output spike to approximate $t^\text{(Ref)}$ of the reference spike.

Here, we summarize the difference between our algorithm and the algorithms reported  before \cite{Mostafa2018supervised, Comsa2019temporal}.
These previous studies were also concerned with the supervised learning of SNNs based on temporal coding.
They used almost the same loss function like an exponential form of spike timing in \cite{Mostafa2018supervised}.
Their approach to avoid the destructively large weight updates was to normalize the gradient when it became excessively large \cite{Mostafa2018supervised},
or to clip the gradient to a fixed value \cite{Comsa2019temporal}.
We use a simple method \eq{suppression1} and \eq{suppression2}, which can prevent the magnitude of the updated weights from becoming too large.
They also reported the spike disappearance problem as explained above.
Mostafa \cite{Mostafa2018supervised} avoided this problem by introducing a penalty term that increased the sum of values of weights and the $L2$ regularization term, but the two terms need to be balanced appropriately.
Comsa {\it et al.} \cite{Comsa2019temporal} also avoided this problem by adding a small penalty that increased the sum of weights when a neuron did not fire.
On the other hand, we avoid this problem by introducing the temporal penalty term, which is a more direct way to stabilize the timing of an output spike.  
Our approach enhances the performance compared to that of Mostafa \cite{Mostafa2018supervised}, and is slightly superior to that of Comsa {\it et al.} \cite{Comsa2019temporal}.

\subsection{Deep learning with SNNs}

Because the important properties of deep learning are sequential nonlinear transformations over layers and end-to-end learning, 
the proposed algorithm can be considered to be a form of deep learning.
The proposed SNN processes information by converting the pattern of an input spike $\bm{t}^{(0)}$ in a layer-by-layer manner as follows:
\begin{flalign}
\bm{t}^{(0)} \rightarrow  \bm{t}^{(1)} \rightarrow \cdots \rightarrow \bm{t}^{(M)} 
\end{flalign}
where $\bm{t}^{(l)} : =  \{t_1^{(l)},t_2^{(l)}, \dots, t_{N^{(l)}}^{(l)}  \}$. 
Note that $t_i^{(l)}$ is considered to be a sufficiently large constant if the $i$th neuron in the $l$th layer does not fire at the end of the simulations.
It is worth noting that the layer-to-layer nonlinear transformation in SNNs is different from that in ANNs.
According to \eq{SpikeTime}, the spike timing $t_i ^{(l)}$ is a linear function of the presynaptic spike timing $t_j ^{(l-1)}$ if $t_i ^{(l)} > t_j ^{(l-1)}$, 
whereas it does not depend on $t_j ^{(l-1)}$ if $t_i ^{(l)} \le t_j ^{(l-1)}$.  
This function seems to be similar to a rectified linear unit (ReLU)\cite{Nair2010rectified}, which is commonly used as an activation function for ANNs.
However, unlike ReLUs, the turning point, $t_i ^{(l)} = t_j ^{(l-1)}$, is determined by other spikes and their weights.
In other words, the information is processed by using the timing order of spikes, which is the most unique characteristic of SNNs based on temporal coding.

\subsection{Consideration of VLSI implementation}

\begin{figure}
	\begin{center}
		\includegraphics[clip, width=8cm]{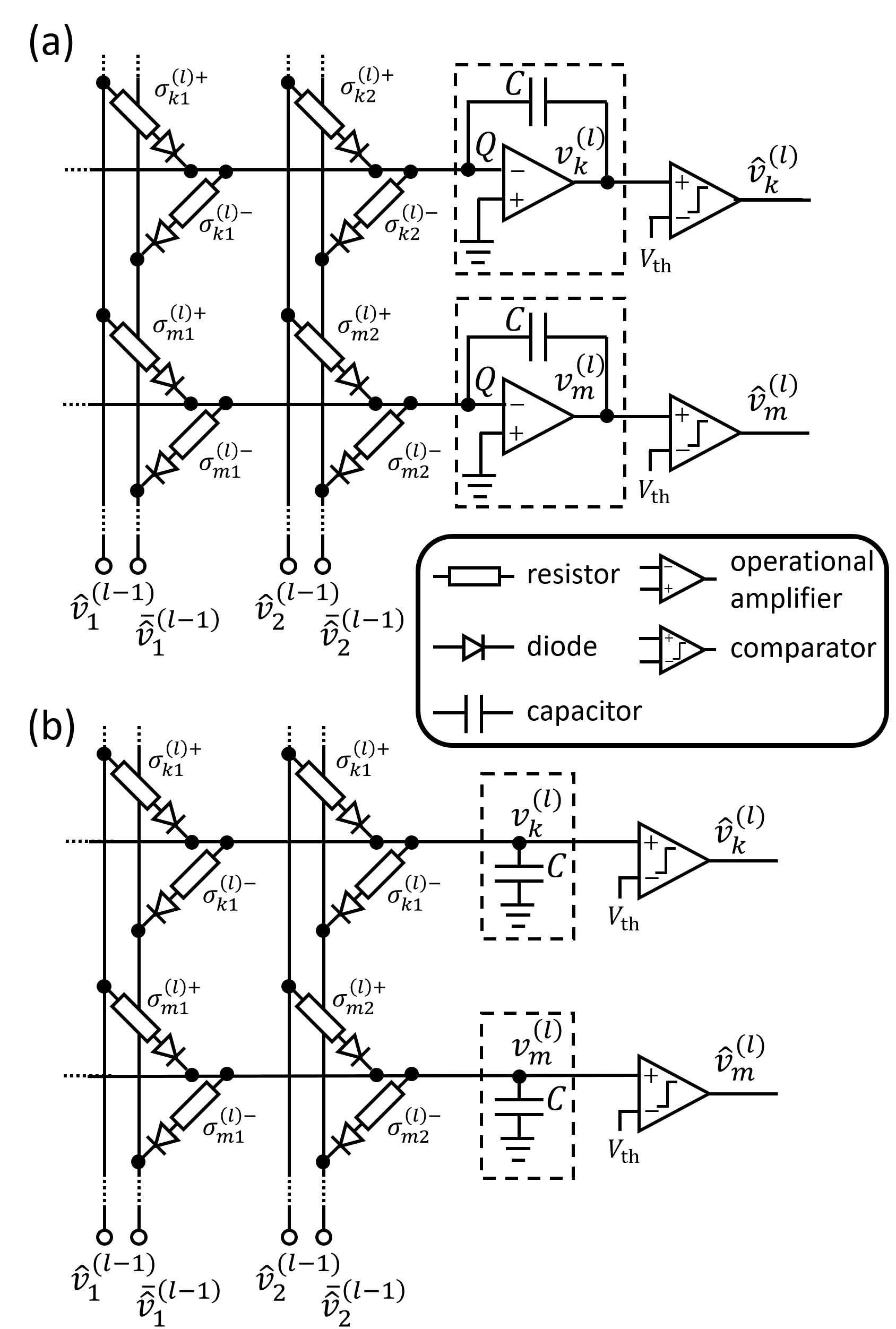}
		\caption{ Examples of implementations of analog circuits of spiking neurons. 
Dashed squares are integrators, which convert summed currents into voltages:
circuit implementation (a) with and (b) without operational amplifiers.
      }
		\label{fig:circuit}
	\end{center}
\end{figure}

For analog VLSI implementations, the dynamics of SNNs must be physically realized in circuits that are carefully designed using either the nonlinear or linear properties of transistors.
Here, we show that the time evolution of the membrane potential of neurons as expressed by \eq{neuron} can be mapped to circuits with analog resistive memory in a simple and straightforward manner.

\fig{circuit}(a) shows an example of such implementations including the analog resistive memory, operational amplifiers, capacitors, and comparators.
The membrane potential $v_i^{(l)}$ is represented as the voltage at the capacitor $C$, which is determined by the charge at the node $Q$.
A spike is generated by a comparator when the membrane voltage exceeds the firing threshold voltage $V_\text{th}$.
The existence of the spike is represented with the output voltage $\hat{v}_i ^{(l)}$ and $\bar{\hat{v}}_i ^{(l)}$ as follows:
\begin{flalign}
\left( \hat{v}_i ^{(l)}(t), \bar{\hat{v}}_i ^{(l)}(t) \right) = \begin{cases}
\left( V_\text{pulse} ^-, V_\text{pulse} ^+ \right),  & ~(t < t_i ^{(l)})\\
\left( V_\text{pulse} ^+, V_\text{pulse} ^- \right), &~(t \ge t_i ^{(l)})
\end{cases} 
\end{flalign}
where $V_\text{pulse}^\pm$ are constants and $V_\text{pulse}^- < 0 < V_\text{th} < V_\text{pulse}^+$ is assumed.

In \fig{circuit}, $\sigma _{ij} ^{(l)+}$ and $\sigma _{ij} ^{(l)-}$ represent the conductance (the inverse of the resistance) from the $j$th neuron in the $(l-1)$th layer to the $i$th neuron in the $l$th layer.
This combination can represent positive weights and negative weights as explained later.
Currents flow through the conductance $\sigma _{ij} ^{(l)\pm}$ to the $i$th neuron in the $l$th layer, 
which causes the membrane potential $v_i ^{(l)}$ to change.
It should be noted that, in \fig{circuit}(a), the voltage at node Q is virtually grounded because of the operational amplifiers.
The time evolution of the membrane potential $v_i ^{(l)}$ is obtained by Kirchhoff's current law and Ohm's law as follows:
\begin{flalign}
C \frac{d}{dt} v_i ^{(l)} (t) = \sum _{j=1} ^{N^{(l-1)}} \left(\sigma _{ij} ^{(l)+} V_\text{pulse}^+ + \sigma _{ij} ^{(l)-} V_\text{pulse}^- \right) \theta(t - t_j ^{(l-1)}). \label{eq:neuron_circuit}
\end{flalign}
By setting $C=1$ and $\sigma _{ij} ^{(l)+} V_\text{pulse}^+ +  \sigma _{ij} ^{(l)-} V_\text{pulse}^- = w_{ij}^{(l)}$, \eq{neuron_circuit} agrees with \eq{neuron}.

We introduce another simpler circuit configuration, as shown in \fig{circuit}(b), 
to allow us to investigate the effect of an approximation of the neuron model \eq{neuron} on the learning performance.
This circuit implementation is similar to that in \cite{Wang2018time}, which was proposed for calculating vector-matrix operations.
Because this circuit (\fig{circuit}(b)) uses no operational amplifiers, which often consume the largest amount of energy in analog circuits,
this implementation is expected to be more energy efficient.
Similar to \eq{neuron_circuit}, the time evolution of the membrane potential $v_i ^{(l)}$ is performed by using Kirchhoff's current law and the Ohm's law as follows:
\begin{flalign}
C\frac{d}{dt} v_i ^{(l)} (t) =& \sum _{j=1} ^{N^{(l-1)}} \sigma _{ij} ^{(l)+} (V_\text{pulse}^+ - v_i^{(l)})\theta(t - t_j ^{(l-1)})\nonumber\\
&+\sigma _{ij} ^{(l)-} (V_\text{pulse}^- - v_i^{(l)})\theta(t - t_j ^{(l-1)}) \nonumber \\
=&\sum _{j=1} ^{N^{(l-1)}}  w_{ij}^{(l)}\delta _{w_{ij}^{(l)}\ge 0} \left(1- \frac{v_i^{(l)}}{V_\text{pulse}^+}\right)\theta(t - t_j ^{(l-1)}) \nonumber\\
&+w_{ij}^{(l)}\delta _{w_{ij}^{(l)}< 0} \left(1- \frac{v_i^{(l)}}{V_\text{pulse}^-}\right)\theta(t - t_j ^{(l-1)}) \nonumber \\
=&\sum _{j=1} ^{N^{(l-1)}}  w_{ij}^{(l)} \left(1 - v_i^{(l)}V_\text{pulse}^{ij}\right)\theta(t - t_j ^{(l-1)}) \label{eq:OpAmpFree}
\end{flalign}
where we define   
\begin{flalign}
\delta _{s} &:= \begin{cases}
0, \text{ if a condition $s$ is false} \\
1, \text{ if a condition $s$ is true}
\end{cases}\\
V_\text{pulse}^{ij} &:= \frac{\delta _{w_{ij}^{(l)}\ge 0}}{V_\text{pulse}^+} + \frac{\delta _{w_{ij}^{(l)}< 0}}{V_\text{pulse}^-}.
\end{flalign}
Note that the following relationships
\begin{flalign}
\sigma _{ij} ^{(l)+} &= \frac{w_{ij}^{(l)}\delta _{w_{ij}^{(l)}\ge 0}}{V_\text{pulse}^+}~~~\sigma _{ij} ^{(l)-} = \frac{w_{ij}^{(l)}\delta _{w_{ij}^{(l)}< 0}}{V_\text{pulse}^-}
\end{flalign}
were used.
By setting $C=1$ and letting $V_\text{pulse}^{ij} \rightarrow 0$, \eq{OpAmpFree} converges to \eq{neuron}, 
where $V_\text{pulse}^{ij} \rightarrow 0$ is realized by $|V_\text{pulse}^\pm| \rightarrow \infty$.
However, smaller values of $|V_\text{pulse}^\pm|$ are more preferable in terms of energy efficiency.
The effects of the finite values of $|V_\text{pulse}^\pm|$ on the learning performance were not previously considered \cite{Wang2018time}.
Thus, we investigated the effects by numerical simulations.
Except when explicitly pointed out, simulations were conducted with the neuron model expressed by \eq{neuron}.
Details of the simulation method and the learning methods for the neuron model expressed by \eq{OpAmpFree} are provided in the appendix.
We note that if current sources (field-effect transistors) are used instead of resistors, as employed by others \cite{Yamaguchi2019energy},
the time evolution equation becomes similar to \eq{neuron_circuit} corresponding to the case of \fig{circuit}(a).

Finally, we explain the advantages of SNNs over time-domain circuits of ANNs\cite{Marinella2018multi, Bavandpour2019energy, Yamaguchi2019energy,Wang2018time} in terms of the energy efficiency of circuits.
First, the circuits of Marinella {\it et al.} \cite{Marinella2018multi} need Analog-to-Digital Converters (ADCs) 
to convert the signal for each layer of ANNs, which are known as energy-hungry processes.
Actually, it is reported that ADCs were the energy bottleneck in the systems \cite{Marinella2018multi}.
Circuit implementations requiring no ADCs were proposed \cite{Bavandpour2019energy, Yamaguchi2019energy,Wang2018time}.
These circuits compute the positive and negative weights independently, and subsequently take the difference of the two.
As a result of the above principles, these circuits require two circuits to compute the positive and negative weights, 
and the dynamic range decreases significantly because the difference is taken (corresponding to digit loss in digital computing).
The advantage of using SNNs over these circuits is that our circuits shown in \fig{circuit} require no full ADCs because the input and output of the circuits consist of binary spikes,
and that the circuits can compute the positive and negative weights at the same time.

\section{Experimental setup} \label{ss:experimental_setup}

In this section, we describe the experimental setup for evaluating the learning performance of the proposed algorithm with the MNIST dataset\cite{Lecun1998gradient}.
We also present the experimental setup for evaluating the robustness of the algorithms against manufacturing variations in the devices, which are unavoidable in analog VLSI implementations.
In all the experiments, we set $V_\text{th}=1$.
We again note that neuron model \eq{neuron} was used unless otherwise noted.

\subsection{Model evaluation with MNIST dataset} \label{ch:setup_MNIST}
The MNIST dataset \cite{Lecun1998gradient} consists of 70,000 images of handwritten digits, 
each of which is represented as a $784~(=28\times28)$-dimensional vector with a label of a corresponding integer (from 0 to 9).
In our experiments, 60,000 images were used for training, and the remaining  10,000 images were used for testing, as in \cite{Lecun1998gradient}.

To process data in SNNs, the vector data are converted into spike patterns as follows:
we normalize all the elements (pixels) of the vector $\bm{x}$ as $0 \le x_i \le 1$. 
Then, the corresponding input spikes are generated at 
\begin{flalign}
t_i ^\text{(0)} = \tau ^\text{(in)}\left(1 - x_i\right), \text{ if } x_i > 0 \label{eq:Encoding_conv} 
\end{flalign}
where $\tau ^\text{(in)}$ is the maximum time window of the input spikes, and was set at $5$ ms in our experiments. 
A spike is not generated when $x_i=0$.
At the training phase, to improve the generalization performance, 
noise was introduced according to a Gaussian distribution $N(0, \sigma _t)$ with zero mean and  standard deviation $\sigma _t$.

\subsection{Robustness against device manufacturing variations} \label{ss:experiments_robustness}

In analog VLSI implementations, variations in the manufactured devices are unavoidable, and must be taken into account during circuit design.
In this experiment, we investigated the effect of manufacturing variations on the learning performance, which has yet not been reported for SNNs based on temporal coding, to the best of our knowledge.
To suppress these effects, we employed a learning technique that is described later.

In the VLSI implementations of \fig{circuit}(a) and \fig{circuit}(b), 
various parameters, such as weights, capacitance, spike delays, and firing threshold, need to be taken into account.
Owing to the computational complexity, we investigated the effects of variations on the learning performance of only the firing threshold and spike delays in this study.
We chose these two parameters because they affect the network dynamics in a different way: 
the firing threshold changes the generation timing of spikes, whereas the spike delays shift the arrival timing of spikes.
The other parameters can be investigated in the same way we explained below.
The remainder of this section describes the mathematical representations of the variations and then the evaluations of the effects of the variations.

First, we mathematically model the variations of the firing threshold $V_\text{th}$.
We define the firing threshold of the $i$th neuron in the $l$th layer as $V_{\text{th}}^{(l,i)}$.
For simplicity, we assume that the probability density distribution of the variation is a clipped Gaussian distribution as follows:
\begin{flalign}
V_{\text{th}}^{(l,i)} &= \max \left( \hat{V}_{\text{th}}^{(l,i)}, 0 \right) \label{eq:Vth_std} \\
\hat{V}_{\text{th}}^{(l,i)} &\sim N(V_\text{th},\sigma _{V_\text{th}})  
\end{flalign}
where $N(V_\text{th},\sigma _{V_\text{th}})$ represents a Gaussian distribution with mean $V_\text{th}$ and standard deviation $\sigma _{V_\text{th}}$ 
and the operation `$\sim$' means drawing a value from the distribution.
\begin{figure}
	\begin{center}
		\includegraphics[clip, width=8cm]{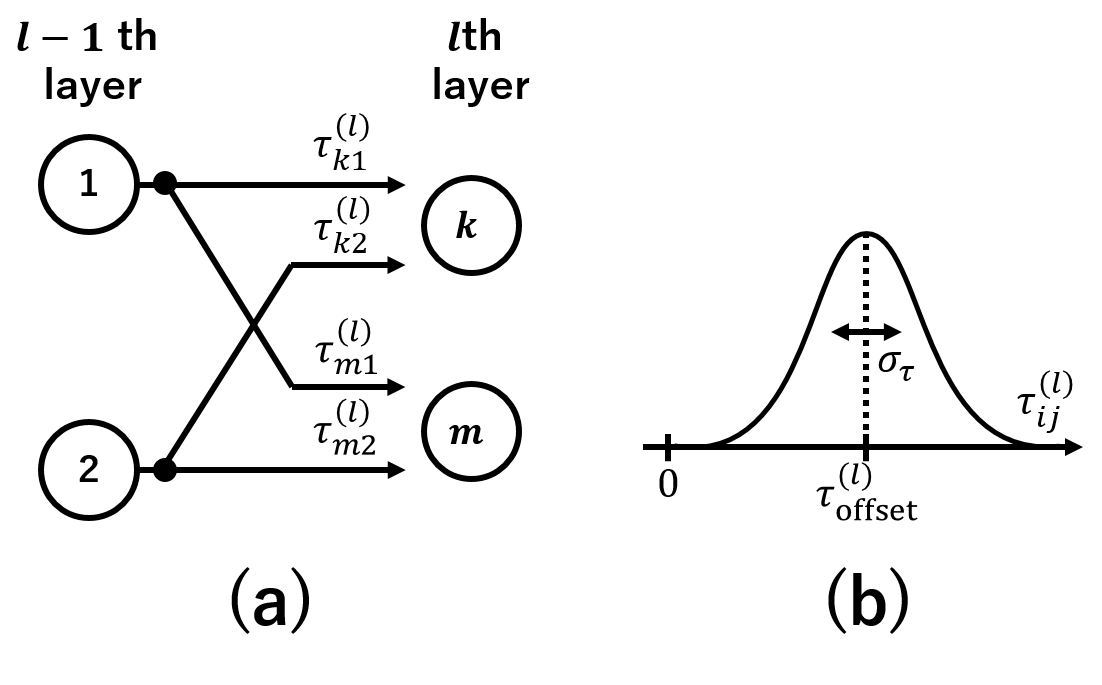}
		\caption{(a) Schematic diagram of the spike delays. Each connection has its own delay such as $\tau _{k1}^{(l)}$. 
(b) The probabilistic distribution of spike delays given by \eq{delay_distribution}. 
Assuming that the mean $\tau _\text{offset}^{(l)}$ of the Gaussian distribution is much larger than the standard deviation $\sigma _\tau$, 
the probability of drawing negative values from \eq{delay_distribution} is negligible.
    }
		\label{fig:delay_model}
	\end{center}
\end{figure}

Second, we mathematically model the variation of spike delays.
As shown in \fig{delay_model}(a), spike delays from the $j$th neuron in the $(l-1)$th layer to the $i$th neuron in the $l$th layer are defined as $\tau _{ij} ^{(l)}$.
We assume a Gaussian probability density distribution of the spike delays
\begin{flalign}
\tau _{ij} ^{(l)} \sim N(\tau _\text{offset} ^{(l)},\sigma _\tau). \label{eq:delay_distribution}
\end{flalign}
We also assume
\begin{equation}
\tau _\text{offset} ^{(l)} \gg \sigma _\tau ~ \forall ~ l 
\end{equation}
such that the condition $\tau _{ij} ^{(l)} > 0 $ is always satisfied as shown in \fig{delay_model} (b). 
The offset $\tau _\text{offset}^{(l)}$ can be removed from the simulation as follows.
Consider the modification of $\tau _{ij} ^{(l)}$ in the SNNs.
If we add a constant $\alpha^{(l)}$ to $\tau _{ij} ^{(l)}$ for all $i$ and $j$, 
all the spike timings of the neurons in the $l$th-layer are delayed by $\alpha^{(l)}$, e.g., 
for the first layer the delays  are $\alpha^{(1)}$ and  
for the second layer the delays are $\alpha^{(1)} + \alpha ^{(2)}$.
Therefore, the spike timing in the output layer is delayed as $\sum _{l=1} ^{M} \alpha^{(l)}$.
We denote the delayed spike timing by $\hat{\bm{t}}^{(l)}$ for the $l$th layer.
Note that the loss function shown in \eq{CCE_Softmax} is invariant against the delays as
\begin{flalign}
L(\bm{t}^{(M)}, \bm{\kappa};\bm{w}) = L(\hat{\bm{t}}^{(M)}, \bm{\kappa};\bm{w})
\end{flalign}
because the effects of the difference $\sum _{l=1} ^{M} \alpha^{(l)}$ between $\bm{t}^{(M)}$ and $\hat{\bm{t}}^{(M)}$ are cancelled out in \eq{softmax}.
Moreover, in \eq{temporal_penalty}, the temporal penalty term $R(\bm{t}^{(M)})$ is invariant against the delays 
if we add the delay $\sum _{l=1} ^{M} \alpha^{(l)}$ to the reference spike timing as 
\begin{flalign}
\hat{t}^\text{(Ref)} = t^\text{(Ref)} + \sum _{l=1} ^{M} \alpha^{(l)}.
\end{flalign}
Therefore, the cost function shown in \eq{Cost} becomes invariant against the delays $\{\alpha^{(l)}\}_{l=1}^M$.
This invariance indicates that a constant shift of the delay $\tau _\text{offset} ^{(l)}$ does not affect the learning results.
Thus, our numerical simulations were conducted with $\tau_\text{offset}^{(l)} = 0$.
Note that a situation in which the spike timing $t_i^{(l)}$ causes the spike to arrive earlier than the arrival time $t_j^{(l-1)} + \tau_{ij}^{(l)}$ may occur in the numerical simulation.
However, information processing in feedforward SNNs is valid because the differences among the arrival times within each layer are maintained.

Using the variation models for the firing threshold $V_\text{th}$ and for spike delays $\tau _{ij} ^{(l)}$,
we investigated the effects of the variations on the learning performance in the following two cases:
(i) the exact values of the parameters ($V_\text{th}$ and $\tau _{ij} ^{(l)}$) are not known, but those distributions are known;
(ii) the exact values of the variation parameters are known.

In case (i), the dynamics in the SNNs is black boxed, thus direct training is not possible.
To overcome this problem, we employed a learning technique in the training phase.
We extract the values of these parameters from \eq{Vth_std} or \eq{delay_distribution} with estimated standard deviations $\sigma _{V_\text{th}(\tau)}^\text{(train)}$. 
The same techniques were reported for training binarized neural networks \cite{Miyashita2017neuro}.
In case (ii), training the SNNs can be carried out in a straightforward manner.

In the test phase, we evaluated the performance of the SNNs with various standard variations $\sigma _{V_\text{th}(\tau)}^\text{(test)}$.
Note that the experimental setup described in section \ref{ch:setup_MNIST} corresponds to the case of $\sigma_{V_\text{th}(\tau)} ^\text{(train)}=\sigma_{V_\text{th}(\tau)} ^\text{(test)}=0$ in case (i).

\section{Results} \label{ss:results}

\subsection{Performance on the  MNIST dataset}

\begin{figure}
	\begin{center}
		\includegraphics[clip, width=9cm]{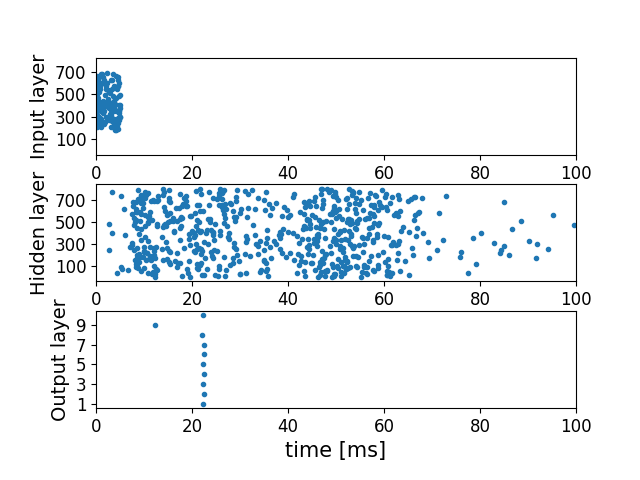}
		\caption{Raster plots of the spike timing of the neurons in the input, hidden, and output layers, from top to bottom, respectively, for a 2-layer SNN (784-800-10).
		The hyperparameters are $\eta=1500, t^\text{(Ref)}=21\text{ ms}, \gamma=100, \text{ and }\epsilon=4$.
    }
		\label{fig:raster_plot}
	\end{center}
\end{figure}

\begin{figure}
	\begin{center}
		\includegraphics[clip, width=9cm]{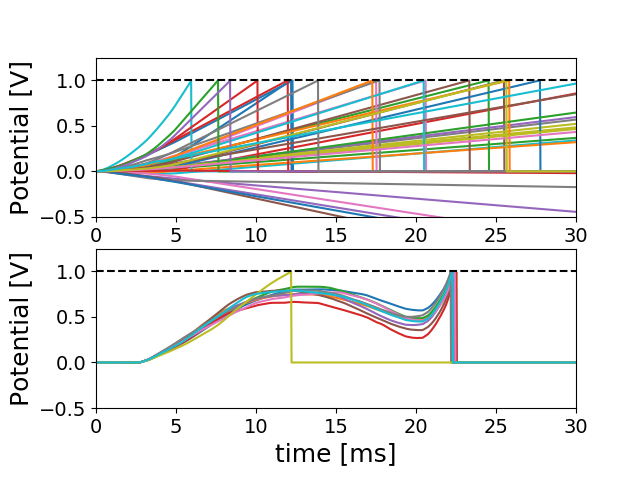}
		\caption{Time evolution of the membrane potentials of the neurons in the hidden layer (top) and the output layer (bottom) for a 2-layer SNN (784-800-10).
    The horizontal dashed lines represent the value of $V_\text{th}~(=1)$. The hyperparameters are $\eta=1500, t^\text{(Ref)}=21\text{ ms}, \gamma=100, \text{ and }\epsilon=4$.
    }
		\label{fig:time_evolution}
	\end{center}
\end{figure}

\fig{raster_plot} shows the spike timing pattern (a raster plot) of a 2-layer SNN (784-800-10) after training for the input, the hidden, and output layers, from top to bottom, respectively.
The horizontal axis represents the time in milliseconds and the vertical axis represents the index of neurons for each layer. 
These results show that the broad distribution of the spike timing in the hidden layer is such that a large portion of neurons in the hidden layer fire after the output neurons fire.
In the output layer, the ninth output neuron fires the earliest, which corresponds to the correct label $(=8)$.
\fig{time_evolution} shows the time evolution of the membrane potential $v_i^{(l)}(t)$ of the neurons in each layer. 
The membrane potentials evolve as \eq{neuron} such that their behavior can be expressed by a piecewise linear function.
For the hidden layer, because the neurons receive no spikes from the input layer after 5 ms, their membrane potentials evolve linearly after the time 5 ms.
For the output neurons, their membrane potential varies nonlinearly because they receive many spikes from the hidden layer.
The membrane potential of the output neurons was observed to be suppressed by spikes from the hidden layer at approximately $10$ ms except that of the output neuron corresponding to the correct label.
One neuron fired at approximately 12 ms and the others at approximately $t^\text{Ref}$. 
These results indicate that our proposed temporal penalty term is effective. 

\begin{figure}
	\begin{center}
		\includegraphics[clip, width=9cm]{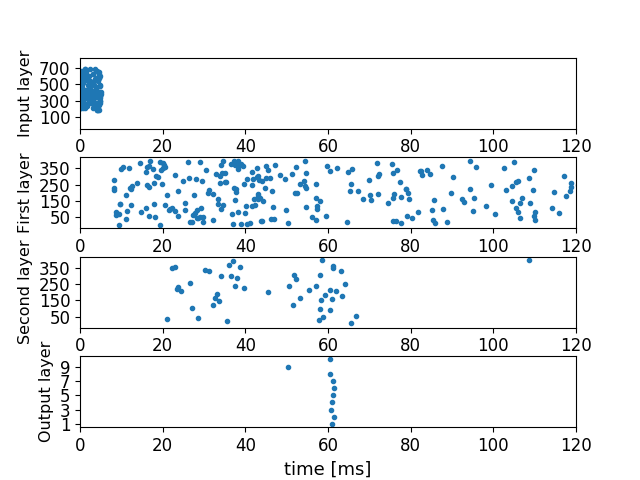}
		\caption{Raster plots for a 3-layer SNN (784-400-400-10). 
		The hyperparameters are $\eta=200, t^\text{(Ref)}=60\text{ ms}, \gamma=100, \text{ and }\epsilon=1$.
      }
		\label{fig:raster_plot_2_hidden}
	\end{center}
\end{figure}

\begin{figure}
	\begin{center}
		\includegraphics[clip, width=9cm]{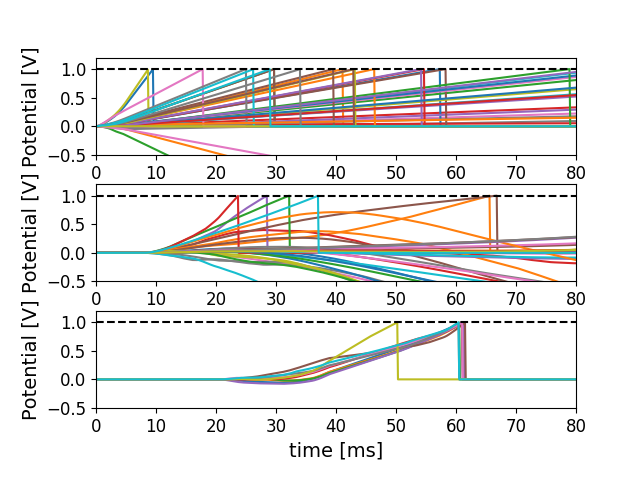}
		\caption{Time evolution of the membrane potentials of the neurons in the first layer (top), in the second layer (middle), and in the output layers (bottom), for a 3-layer SNN (784-400-400-10). The horizontal dashed lines represent the values of $V_\text{th}(=1)$. The hyperparameters are $\eta=1500, t^\text{(Ref)}=60\text{ ms}, \gamma=100, \text{ and }\epsilon=1$.
      }
		\label{fig:time_evolution_2_hidden}
	\end{center}
\end{figure}

Similar but slightly different network dynamics is observed for 3-layer SNNs (784-400-400-10).
\fig{raster_plot_2_hidden} shows the raster plots of the spikes in the SNN. 
The distribution of spikes in the first hidden layer was broad, 
whereas that in the second hidden layer was relatively narrow.
A large portion of neurons in the second hidden layer are observed not to have fired.
This reduction in the number of firing neurons may imply that the neurons needed for information processing are selected for each data.
\fig{time_evolution_2_hidden} shows the time evolution of the membrane potential of the neurons in each layer. 
The results show that the membrane potentials of neurons in the second and output layers evolve nonlinearly, 
whereas those in the first layer evolve linearly after 5 ms.

\begin{table*} 
\begin{center}
\caption{Performance comparisons on the MNIST dataset}
\begin{tabular}{lcl}
Network &  Coding &  Accuracy \\ \hline \hline
784-800-10 \cite{Srivastava2014dropout} & ANN &  98.4 \% \cite{Srivastava2014dropout} \\ 
784-800-800-10 \cite{Wan2013regu} & ANN &  98.8 \% \cite{Wan2013regu} \\ \hline
784-800-10 \cite{Lee2016training} & SNN (rate coding) & 98.64 \% \cite{Lee2016training} \\ 
784-300-300-10 \cite{Lee2016training} & SNN (rate coding) & 98.77 \% \cite{Lee2016training} \\  \hline
784-800-10 \cite{Mostafa2018supervised} & SNN (temporal coding) &  97.55 \% \cite{Mostafa2018supervised} \\ 
784-400-400-10 \cite{Mostafa2018supervised} & SNN (temporal coding)& 97.14 \% \cite{Mostafa2018supervised} \\ 
784-340-10 \cite{Comsa2019temporal}& SNN (temporal coding) & 97.96\% \cite{Comsa2019temporal} \\
784-500-10 (This work) & SNN (temporal coding)&  97.83$\pm$0.07 \%\\   
784-800-10 (This work) & SNN (temporal coding)&  {\bf98.04$\pm$0.05} \% \\ 
784-300-300-10 (This work) & SNN (temporal coding)& 97.73$\pm$0.08 \% \\
784-400-400-10 (This work) & SNN (temporal coding)& 97.90$\pm$0.12 \%  
\end{tabular}
\label{tab:MNIST_accuracy}
\end{center}
\end{table*}

\tab{MNIST_accuracy} shows the results of the classification accuracy on the MNIST dataset.
The standard errors `$\pm$' for our results were obtained from 10 trials with different initial weights.
The SNNs based on rate-coding \cite{Lee2016training} performed as good as ANNs\cite{Srivastava2014dropout,Wan2013regu}, 
whereas the SNNs based on temporal coding \cite{Mostafa2018supervised} performed slightly worse than ANNs.
However, the improvements in our learning algorithm increased the performance of our algorithm compared with previous research \cite{Mostafa2018supervised}, 
and the performance was slightly higher than \cite{Comsa2019temporal}.
Our best result is highlighted in bold in \tab{MNIST_accuracy}.

\begin{figure}
	\begin{center}
		\includegraphics[clip, width=9cm]{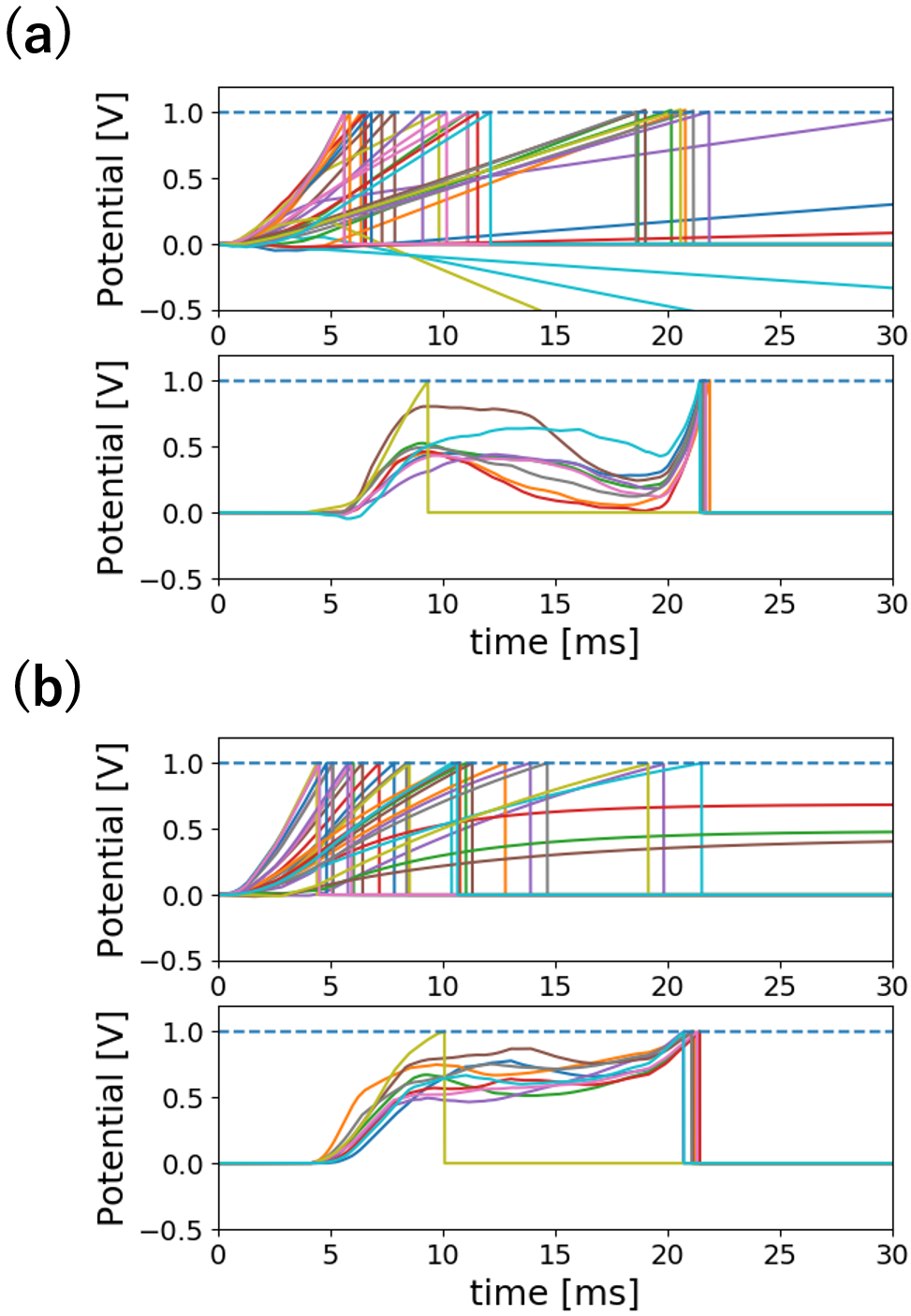}
		\caption{Time evolution of the membrane potentials of the neurons for 2-layer SNN (169-300-10) for $V_\text{pulse}$. (a) $V_\text{pulse}=128$, (b) $V_\text{pulse}=2$. 
		In both (a) and (b), the upper and lower panels represent the first and the output layer, respectively. 
		In the upper panel of (b), thee membrane potentials converge to some intermediate constants.	
		The hyperparameters are $\eta=1500, t^\text{(Ref)}=21\text{ ms}, \gamma=8, \text{ and } \epsilon=10$.
      }
		\label{fig:time_evolution_Vpulse}
	\end{center}
\end{figure}

\begin{figure}
	\begin{center}
		\includegraphics[clip, width=9cm]{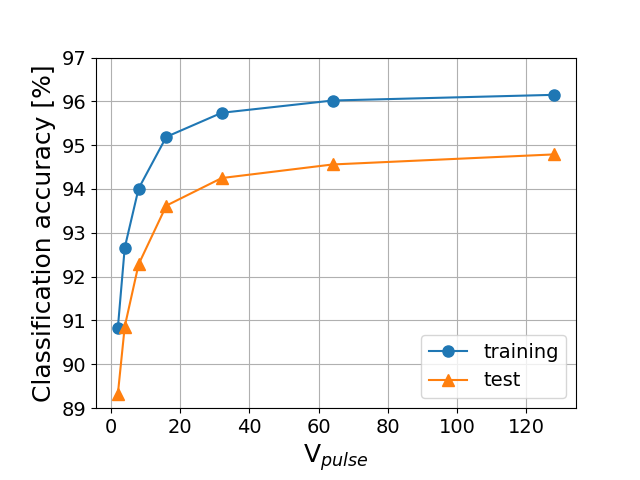}
		\caption{Classification accuracy on the shrunk MNIST dataset with various values of $V_\text{pulse}$ ($=2, 4,8, \dots ,128$) for 2-layer SNN (169-300-10). Each plot is the average of the results of five trials. 
		The standard errors are smaller than the symbol size.  
		The hyperparameters are $\eta=1500, t^\text{(Ref)}=21\text{ ms}, \gamma=8, \text{ and }\epsilon=10$.
    }
		\label{fig:MNIST_Vpulse}
	\end{center}
\end{figure}

Next, we show the results of the learning performance when we replace the original neuron model \eq{neuron} with the alternative neuron model \eq{OpAmpFree}.
We again note that the model \eq{OpAmpFree} is considered to be more energy efficient.
To reduce the computational costs, we focused on a small-scale network (169-300-10) and used a shrunk version of the MNIST dataset.
A shrunk MNIST image $\hat{\bm{x}}$ was obtained by convoluting a $4\times4$ kernel 
of which the elements are all $1$ onto an original $28\times28$ image $\bm{x}^*$ with a stride of 2. 
The $(i,j)$ pixel of the shrunk image $\hat{x}_{ij}$ is obtained by
\begin{flalign}
\hat{x}_{ij} &= \frac{1}{16}\sum _{k=0}^{3}\sum _{m=0}^{3}x_{2i+k-1~2j+m-1}^*,~(i,j= 1,2,\dots, 13).
\end{flalign}
Then, we obtained spike patterns by using \eq{Encoding_conv} with the flatted shrunk two-dimensional data $\hat{x}_{ij}$.
In the experiment, we experienced difficulties in achieving learning convergence because of destructively large weight updates 
which had their origins in the temporal penalty term \eq{temporal_penalty}.
We circumvented this difficulty by replacing the temporal penalty term \eq{temporal_penalty} with
\begin{flalign}
R(\bm{t}^{(M)}) &:=   \sum_{i=1} ^{N^{(M)}}\left( t_i ^{(M)} - t^\text{(Ref)} \right)^{3/2}
\end{flalign} 
to soften the effect of its large weight updates when the timing of the output spike is far from $t^\text{(Ref)}$.

\fig{time_evolution_Vpulse} shows the time evolution of the membrane potentials of the neurons for different values of $V_\text{pulse}$ ($=|V_\text{pulse}^+|=|V_\text{pulse}^-|$).
When $V_\text{pulse}=128$ (\fig{time_evolution_Vpulse}(a)), the membrane potentials of the neurons in the hidden layer evolve linearly after 5 ms 
and those in the output layer evolve nonlinearly, which is similar to the case of \fig{time_evolution}.
On the other hand, when $V_\text{pulse}=2$ (\fig{time_evolution_Vpulse}(b)), 
the membrane potentials of the neurons in the hidden evolve nonlinearly, which is caused by the term $v_i^{(l)}V_\text{pulse}^{ij}$ on the right-hand side of \eq{OpAmpFree}.
Moreover, the membrane potentials of the neurons are stabilized at some intermediate value, 
because \eq{OpAmpFree} has an equilibrium point with
\begin{flalign}
v_i^{(l)} = \frac{\sum _{j\in \Gamma _i ^{(l)}} w_{ij}^{(l)} }{\sum _{j\in \Gamma _i ^{(l)}} w_{ij}^{(l)} V_\text{pulse}^{ij}}.
\end{flalign}

\fig{MNIST_Vpulse} shows the results of the classification accuracy for various values of $V_\text{pulse}$ ($=2, 4,8, \dots ,128$).
As the value of $V_\text{pulse}$ decreases, the classification accuracy decreases monotonically.
As shown in \fig{time_evolution_Vpulse}, there exists an equilibrium point in the membrane potential lower than $V_\text{th}$ when the value of $V_\text{pulse}$ is small, 
which may complicate the training.

\subsection{Effects of manufacturing variations on learning performance}

\begin{figure}
	\begin{center}
		\includegraphics[clip, width=9cm]{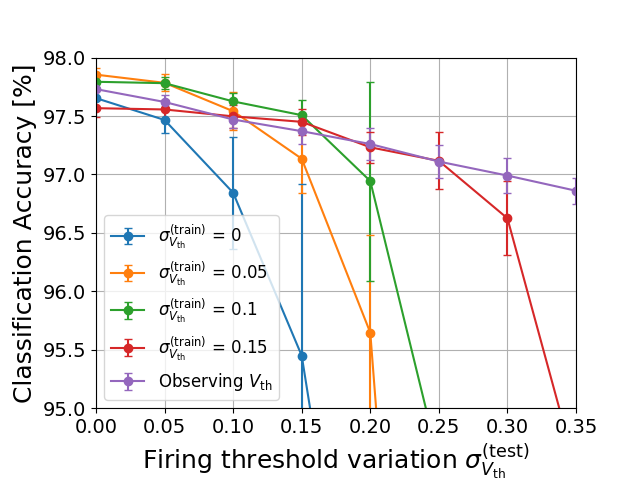}
		\caption{Classification accuracy of 2-layer SNN (784-500-10) with various variations of the firing threshold $\sigma _{V_\text{th}} ^\text{(train)}$ at the training phase. 
		The horizontal axis represents the value of $\sigma _{V_\text{th}} ^\text{(test)}$ at the test phase.	 
		Each plotted point and error bar are the average and standard deviation of the results of ten trials, respectively.
		The hyperparameters are $\eta=1500, t^\text{(Ref)}=21\text{ ms}, \gamma=100, \text{ and }\epsilon=1$.
      }
		\label{fig:Vth_effects}
	\end{center}
\end{figure}

We present the results of the numerical simulation of the experiments described in \ref{ss:experiments_robustness}.
\fig{Vth_effects} shows the results of the classification accuracy 
for various values of the standard deviations of the firing threshold in the test phase $\sigma _{V_\text{th}} ^\text{(test)}$.
When no variation is assumed at the training phase ($\sigma _{V_\text{th}} ^\text{(train)} = 0$), the classification accuracy decreases conspicuously in the test phase as the standard deviation $\sigma _{V_\text{th}} ^\text{(test)}$ increases.
When variations are assumed in the training phase ($\sigma _{V_\text{th}}^\text{(train)} = 0.05,~ 0.1, ~0.15$) corresponding to case (i) in \ref{ss:experiments_robustness}, 
the degradation of the classification accuracy can be suppressed to an extent.
Especially, the suppression is effective when the variation in the test phase is almost the same as that assumed in the training phase ($\sigma _{V_\text{th}}^\text{(train)} \fallingdotseq \sigma _{V_\text{th}} ^\text{(test)}$).
When the values of all $V_\text{th}$ are known corresponding to case (ii) in \ref{ss:experiments_robustness}, 
the degradation of the classification accuracy can be well suppressed.

\begin{figure}
	\begin{center}
		\includegraphics[clip, width=9cm]{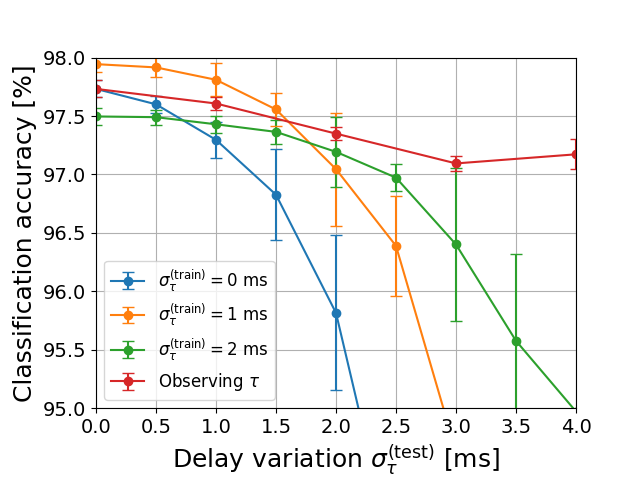}
		\caption{Classification accuracy of 2-layer SNN (784-500-10) with various variations of spike delays $\sigma _\tau ^\text{(train)}$ at the training phase. 
		The horizontal axis represents the value of standard deviation of $\sigma _\tau ^\text{(test)}$ at the test phase.	 
		Each plot and error bar are the average  and standard deviation of the results of ten trials, respectively. 
		The hyper parameters are $\eta=1500, t^\text{(Ref)}=21\text{ ms}, \gamma=100, \text{ and }\epsilon=10$.}
		\label{fig:delay_effects}
	\end{center}
\end{figure}

Next, the results of classification accuracy for various standard deviations of spike delays at the test phase $\sigma _\tau ^\text{(test)}$ are shown in \fig{delay_effects}.
As in the case of variations of firing threshold $V_\text{th}$ in \fig{Vth_effects}, when no variation is assumed at the training phase ($\sigma _\tau ^\text{(train)} = 0$), 
the classification accuracy decreases as the standard deviation $\sigma _\tau ^\text{(test)}$ at the test phase increases.
While, when variations are assumed at the training phase ($\sigma _{\tau}^\text{(train)} = 0,~ 1\text{ ms}, ~2\text{ ms}$) corresponding to case (ii) in \ref{ss:experiments_robustness}, the degradation of classification accuracy can be suppressed to some extent.
Especially, the suppression efficiently works when the variation at the test phase is almost the same as that assumed at the training phase ($\sigma _{\tau}^\text{(train)} \fallingdotseq \sigma _{\tau} ^\text{(test)}$). 
When values of all $\tau^{(l)}_{ij}$ are known corresponding to case (ii) in \ref{ss:experiments_robustness}, the degradation of classification accuracy can be well suppressed.

The series of numerical simulations shown above indicates that our proposed algorithm is robust against the device manufacturing variations.
This fact paved a way to manufacturing VLSIs implementing the proposed algorithm.

\section{Concluding remarks} \label{ss:conclusion}

In this study, we pursued the goal of achieving both high algorithmic performance and high energy efficiency 
by proposing a novel supervised learning algorithm for multilayer SNNs which can be implemented in VLSI circuits with analog resistive memory.
We also proposed novel learning techniques such as adding a temporal penalty term to improve the learning performance.

Although the performance of the proposed algorithm on the MNIST dataset was slightly worse than SNNs based on rate coding and ANNs, 
the performance was confirmed to be as high as the state-of-the-art SNNs based on temporal coding.
We showed that VLSI circuits can be designed with analog resistive memory in a simple and straightforward manner.
We used numerical simulation to investigate the extent to which the learning performance depended on the circuit design.
We also showed that the effects of manufacturing variations, which are unavoidable for real VLSI implementations, 
can be suppressed by estimating the magnitude of variations or by observing the exact values of these parameters.
Therefore, we achieved our goal to a certain extent.
The next step will be the fabrication of VLSI chips based on the circuits and algorithms proposed in this paper.

In this study, we focused on the development of energy-efficient algorithms for feedforward networks as a first step. 
Recently, it was shown that SNNs with recurrent connections can process time-series data with high performance \cite{Nicola2017supervised, Panda2018learning, Bellec2018long}. 
For various applications of edge computing, the development of energy-efficient and hardware-friendly learning algorithms for SNN with recurrent connections is strongly anticipated. 
Thus, the development of these algorithms should be the next goal. 
Finally, we would like to remark that recent advances in biologically plausible and spike-based BP \cite{Lillicrap2016random, Nokland2016direct, Guerguiev2017towards, Neftci2018event, Bellec2019biologically, WHITTINGTON2019theory} are foreseen to be beneficial to construct energy-efficient systems including the training process \cite{Park2019neuro}.

\ifarxiv
\section*{Appendix: Back propagation for case \eq{OpAmpFree}}
\else
\appendix[Back propagation for case \eq{OpAmpFree}] 
\fi

\begin{figure}
	\begin{center}
		\includegraphics[clip, width=8cm]{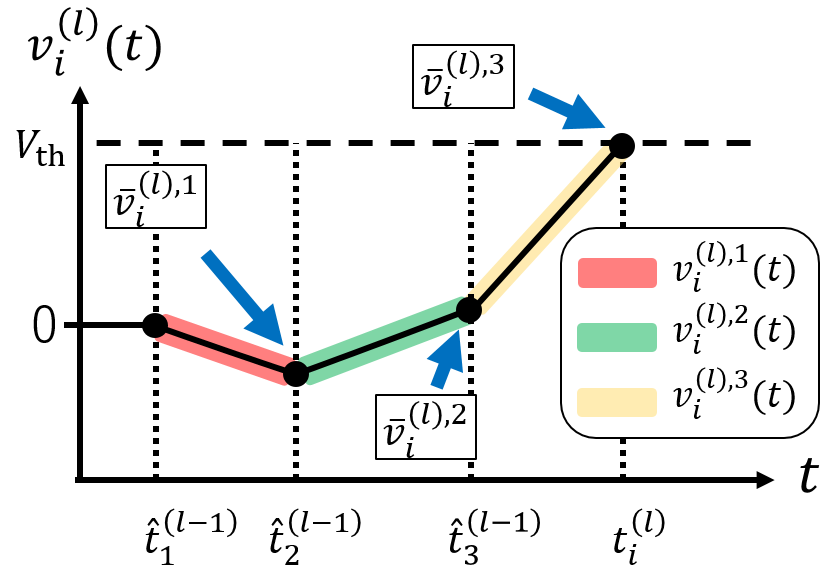}
		\caption{Schematic diagram of the time evolution of the membrane potential.}
		\label{fig:model_RC}
	\end{center}
\end{figure}

Here, we redefine the indices of the neurons to satisfy their indices to be sorted in the order of their spike time as
\begin{flalign}
\hat{t}_1^{(l-1)} \le \hat{t}_2^{(l-1)} \le \dots \le \hat{t}_{N^{(l-1)}}^{(l-1)}. \label{eq:sorted_t} 
\end{flalign}
Then, we obtain the corresponding sorted weights
\begin{flalign}
\hat{w}_{i1}^{(l)}, \hat{w}_{i2}^{(l)}, \dots, \hat{w}_{iN^{(l)}}^{(l)}. \label{eq:sorted_w}
\end{flalign}

The time evolutions of the membrane potentials are piecewise 
such that they can be represented as the corresponding equations.
For the $i$th neuron in the $l$th layer, from the moment when it receives the $k$th earliest spike to when it receives the $(k+1)$th earliest spike, 
the time evolution of its membrane potential $v_i^{(l),k}$ is described as
\begin{flalign}
\frac{d}{dt} v_i ^{(l),k}&(t) = \sum _{j=1}^{k}   \hat{w}_{ij}^{(l)} \left(1 - v_i^{(l),k} V_\text{pulse}^{ij}\right),  \\
\text{ for }& \hat{t}_k^{(l-1)} \le t < \hat{t}_{k+1}^{(l-1)}, ~(k=1,2,\dots ,N^{(l-1)}).
\end{flalign}
The terminal values of $v_i^{(l),k}$ are denoted by $\bar{v}_i^{(l),k}$ as follows:
\begin{flalign}
\bar{v}_i^{(l),k} &:= \begin{cases}
v_i^{(l),k}\left(\hat{t}_{k+1}^{(l-1)}\right), \text{ for } k<G_i^{(l)} \\
v_i^{(l),G}\left(t_i^{(l)}\right), \text{ for } k=G_i^{(l)}
\end{cases} 
\end{flalign}
where $G_i^{(l)}$ represents the number of received spikes before the $i$th neuron fires, which can take different values for different neurons.
Using the above definitions, we can obtain an analytical form of the membrane potential as follows:
\begin{flalign}
\bar{v}_i^{(l),k} &= \begin{cases}
\frac{A_{i1}^{(l)}}{B_{i1}^{(l)}} - \frac{A_{i1}^{(l)}}{B_{i1}^{(l)}}\exp \left(-B_{i1}^{(l)} \Delta \hat{t}_1^{(l)} \right), \text{ for } k=1\\
\frac{A_{ik}^{(l)}}{B_{ik}^{(l)}} - \left(\frac{A_{ik}^{(l)}}{B_{ik}^{(l)}}-\bar{v}_i^{(l),k-1} \right)\exp \left(-B_{ik}^{(l)}\Delta \hat{t}_k^{(l)}\right), \\
~~~~~~~~~~~~~~~~~~\text{ for } k=2,3, \dots, G_i^{(l)}
\end{cases}
\end{flalign}
where 
\begin{flalign}
A_{ik}^{(l)}&:=\sum_{j=1}^k \hat{w}_{ij}^{(l)}~~~B_{ik}^{(l)}:=\sum_{j=1}^k \hat{w}_{ij}^{(l)}V_\text{pulse}^{ij}  \\
\Delta \hat{t}_k^{(l-1)} &:= \begin{cases}
\hat{t}_{k+1}^{(l-1)}-\hat{t}_k^{(l-1)}, \text{ for } k = 1,2,\dots, G_i^{(l)}-1 \\
t_i^{(l)}-\hat{t}_k^{(l-1)}, \text{ for }k=G_i^{(l)}.
\end{cases}
\end{flalign}
Furthermore, the spike timing of the $i$th neuron in the $l$th layer is obtained as follows:
\begin{flalign} 
t_i^{(l)} &= \begin{cases}
\hat{t}_{G_i^{(l)}}^{(l-1)} - \frac{1}{B_{iG_i^{(l)}}^{(l)}} \ln \left(1 -  \frac{B_{iG_i^{(l)}}^{(l)}}{A_{iG_i^{(l)}}^{(l)}}V_\text{th}\right)  , \text{ for } G_i^{(l)}=1 \\
\hat{t}_{G_i^{(l)}}^{(l-1)} - \frac{1}{B_{iG_i^{(l)}}^{(l)}} \ln \frac{ \frac{A_{iG_i^{(l)}}^{(l)}}{B_{iG_i^{(l)}}^{(l)}}-V_\text{th} }{\left(\frac{A_{iG_i^{(l)}}^{(l)}}{B_{iG_i^{(l)}}^{(l)}}-\bar{v}_i^{(l),G_i^{(l)}-1} \right)}, \text{ for } G_i^{(l)}>1.
\end{cases}
\end{flalign}
Fig. \ref{fig:model_RC} shows a schematic diagram of the time evolution of the membrane potential $v_i ^{(l),k}(t)$ .

Similar to \eq{SGD}--\eq{Denom3}, the back propagation rule is then obtained by
\begin{flalign}
\frac{\partial C}{\partial w_{jk} ^{(l)}} &= \frac{\partial \bar{v}_j^{(l),G_i^{(l)}}}{\partial w_{jk}^{(l)}} \delta _j ^{(l)} \label{eq:SGD_circuits} \\
\delta _s ^{(h)} &= \begin{cases}
\frac{\partial t_s^{(h)}}{\partial \bar{v}_s^{(h),G_i^{(l)}}} \sum _i^{N^{(h+1)}} \frac{\partial \bar{v}_i ^{(h+1),G_i^{(l)}}}{\partial t_s^{(h)}} \delta _i ^{(h+1)},  \\
~~~~~~~~~~~~~~~~~\text{ for }h=1,2,\dots, M-1\\
\frac{\partial t_s^{(M)}}{\partial \bar{v}_s^{(M),G_i^{(l)}}} \frac{\partial C}{\partial t_s^{(M)}}, \text{ for } h=M
\end{cases} \label{eq:delta_circuits}\\
\frac{\partial C}{\partial t_i^{(M)}} &=  \kappa_i - S_i +\gamma \left( t_i ^{(M)} - t^\text{(Ref)} \right).
\end{flalign}
Replacing $\{t_i^{(l)}\}$ and $\{w_{ij}^{(l)}\}$ with the sorted expressions $\{\hat{t}_i^{(l)}\}$ and $\{\hat{w}_{ij}^{(l)}\}$ given by \eq{sorted_t} and \eq{sorted_w}, 
the derivatives on the right-hand sides of the above equations are obtained as follows: 

for \eq{SGD_circuits}, 
\begin{flalign}
\frac{\partial \bar{v}_i^{(l),G_i^{(l)}}}{\partial \hat{w}_{ij}^{(l)}} &= \begin{cases}
\sum _{P=j}^{G_i^{(l)}}\frac{\partial \bar{v}_i^{(l),P}}{\partial \hat{w}_{ij}^{(l)}} \delta _{P}^{v_i^{(l)}}, \text{ for }j=1,2,\dots, G_i^{(l)} \\
0, \text{ for }j=G_i^{(l)}+1, G_i^{(l)}+2, \dots, N^{(l-1)}.\\
\end{cases} \label{eq:dvdw_OpAmpFree} 
\end{flalign}
In addition, for \eq{delta_circuits},
\begin{flalign}
\delta _{j}^{v_i^{(l)}} &= \begin{cases}
\prod _{q=j}^{G_i^{(l)}-1} \frac{\partial \bar{v}_i^{(l),q+1}}{\partial \bar{v}_i^{(l),q}} = \frac{\partial \bar{v}_i^{(l),j+1}}{\partial \bar{v}_i^{(l),j}} \delta _{j+1}^{v_i^{(l)}} \\
~~~~~~~~\text{ for } j = 1,2, \dots, G_i^{(l)}-1  \label{eq:delta_v_OpAmpFree} \\
  1, \text{ for }j=G_i^{(l)} \\
\end{cases} \\
\frac{\partial \bar{v}_i ^{(l),G_i^{(l)}}}{\partial \hat{t}_{j}^{(l-1)}} &= \begin{cases}
\frac{\partial \bar{v}_i ^{(l),j}}{\partial \hat{t}_{j}^{(l-1)}} \delta _j^{v_i^{(l)}} \label{eq:dvdt_OpAmpFree}, \text{ for } j=1,2,\dots, G_i^{(l)} \\
0, \text{ for }j=G_i^{(l)}+1, G_i^{(l)}+2, \dots, N^{(l-1)}\\
\end{cases}\\
\frac{\partial t_i^{(l)}}{\partial \bar{v}_i^{(l)}} &= -\left( \frac{\partial \bar{v}_i^{(l),G_i^{(l)}}}{\partial t} \right)^{-1} \nonumber\\
&=\begin{cases}
\frac{-1}{A_{iG_i^{(l)}}^{(l)} - B_{iG_i^{(l)}}^{(l)}V_\text{th}}, \text{ when }G_i^{(l)}\ge 1 \\
\frac{-1}{A_{iG_i^{(l)}}^{(l)} - B_{iG_i^{(l)}}^{(l)}V_\text{th}}, \text{ when } G_i^{(l)}=0.
\end{cases} \label{eq:dtdv_OpAmpFree}
\end{flalign}
Note that the above equations take the value of $0$ when the $i$th neuron does not fire and \eq{dtdv_OpAmpFree} is derived from SpikeProp\cite{Bohte2002error}.
For the calculation of (\ref{eq:dvdw_OpAmpFree}), we obtain the derivatives as follows:
\begin{flalign}
\frac{\partial \bar{v}_i^{(l),k}}{\partial \hat{w}_{ij}^{(l)}} &= \begin{cases}
D_{ij}^{(l),k} \left(1 - \exp \left(-B_{ik}^{(l)}\Delta \hat{t}_k^{(l-1)}\right)\right) \\
~~+ \left(\frac{A_{ik}^{(l)}}{B_{ik}^{(l)}}-\bar{v}_i^{(l),k-1} \right)\Delta \hat{t}_k^{(l-1)}V_\text{pulse}^{ij} \\
~~~~\times \exp \left(-B_{ik}^{(l)}\Delta \hat{t}_k^{(l-1)}\right),\\
~~~~~~~~~~~~~~~~~~~~~\text{ for } k=1,2,\dots,G_i^{(l)} \\
D_{ij}^{(l),k} \left(1 - \exp \left(-B_{ik}^{(l)}\Delta \hat{t}_k^{(l-1)}\right)\right) \\
~~+ \frac{A_{ik}^{(l)}}{B_{ik}^{(l)}}\Delta \hat{t}_k^{(l-1)}V_\text{pulse}^{ij} \\
~~~~\times \exp \left(-B_{ik}^{(l)}\Delta \hat{t}_k^{(l-1)}\right),\\
~~~~~~~~~~~~~~~~~~~~~\text{ for } k=0\\
\end{cases}
\end{flalign}
where $D_{ij}^{(l),k}$ is defined as
\begin{flalign}
D_{ij}^{(l),k} := \frac{\partial }{\partial \hat{w}_{ij}^{(l)}} \left( \frac{A_k^{(l)}}{B_k^{(l)}}\right) &=\frac{B_k^{(l)}-A_k^{(l)}V_\text{pulse}^{ij}}{B_k^{(l)2}}, \nonumber\\
&~~~~~~\text{for } j=1,2,\dots, k.
\end{flalign}
Further, for (\ref{eq:delta_v_OpAmpFree}),
\begin{flalign}
\frac{\partial \bar{v}_i^{(l),k+1}}{\partial \bar{v}_i^{(l),k}} &= \exp \left(-\Delta \hat{t}_{k+1}^{(l-1)} B_{k+1}^{(l)}\right), \nonumber\\ 
&~~~~~~\text{for } k=1,2,\dots, G_i^{(l)}-1.
\end{flalign}
Furthermore, (\ref{eq:dvdt_OpAmpFree}) and (\ref{eq:dtdv_OpAmpFree}) can be calculated with
\begin{flalign}
\frac{\partial \bar{v}_i ^{(l),k}}{\partial \hat{t}_{k}^{(l-1)}} &= \begin{cases}
\frac{\partial \bar{v}_i ^{(l),k-1}}{\partial \hat{t}_{k}^{(l-1)}} \exp \left(-B_{ik}^{(l)} \Delta \hat{t}_k^{(l-1)}\right) \\
~~- \left(A_{ik}^{(l)}-B_{ik}^{(l)}\bar{v}_i^{(l),k-1}\right) \\
~~~~\times \exp \left(-B_{ik}^{(l)}\Delta \hat{t}_k^{(l-1)}\right),\\
~~~~~~~~\text{ for } k=1,2,\dots, G \\
- A_{i0}^{(l)}\exp \left(-B_{i0}^{(l)}\Delta \hat{t}_0^{(l-1)}\right), \\
~~~~~~~~\text{ for }k=0
\end{cases}\\
\frac{\partial \bar{v}_i ^{(l),k-1}}{\partial \hat{t}_{k}^{(l-1)}} &= \begin{cases}
 \left(A_{ik-1}^{(l)}-B_{ik-1}^{(l)}\bar{v}_i^{(l),k-2} \right) \\
 ~~\times \exp \left(-B_{ik-1}^{(l)}\Delta \hat{t}_{k-1}^{(l-1)}\right),\\
 ~~~~~~~~~~~\text{ for }k=2,\dots, G\\
A_{i0}^{(l)}\exp \left(-B_{i0}^{(l)}\Delta \hat{t}_0^{(l-1)}\right),\\
 ~~~~~~~~~~~\text{ for }k=1.
\end{cases} 
\end{flalign}
As described above, the BP for case \eq{OpAmpFree} is obtained.


%

%
%
%

\section*{Acknowledgment}
This work is partially supported by the ``Brain-Morphic AI to Resolve Social Issues'' project and by NEC Corporation.

\ifarxiv
\else
\ifCLASSOPTIONcaptionsoff
  \newpage
\fi




%

\ifarxiv
\bibliographystyle{junsrt}
\else
\bibliographystyle{IEEEtran}
\fi
\bibliography{myBib}

%
%

%

\ifarxiv
\else
\begin{IEEEbiography}[{\includegraphics[width=1in,height=1.25in,clip,keepaspectratio]{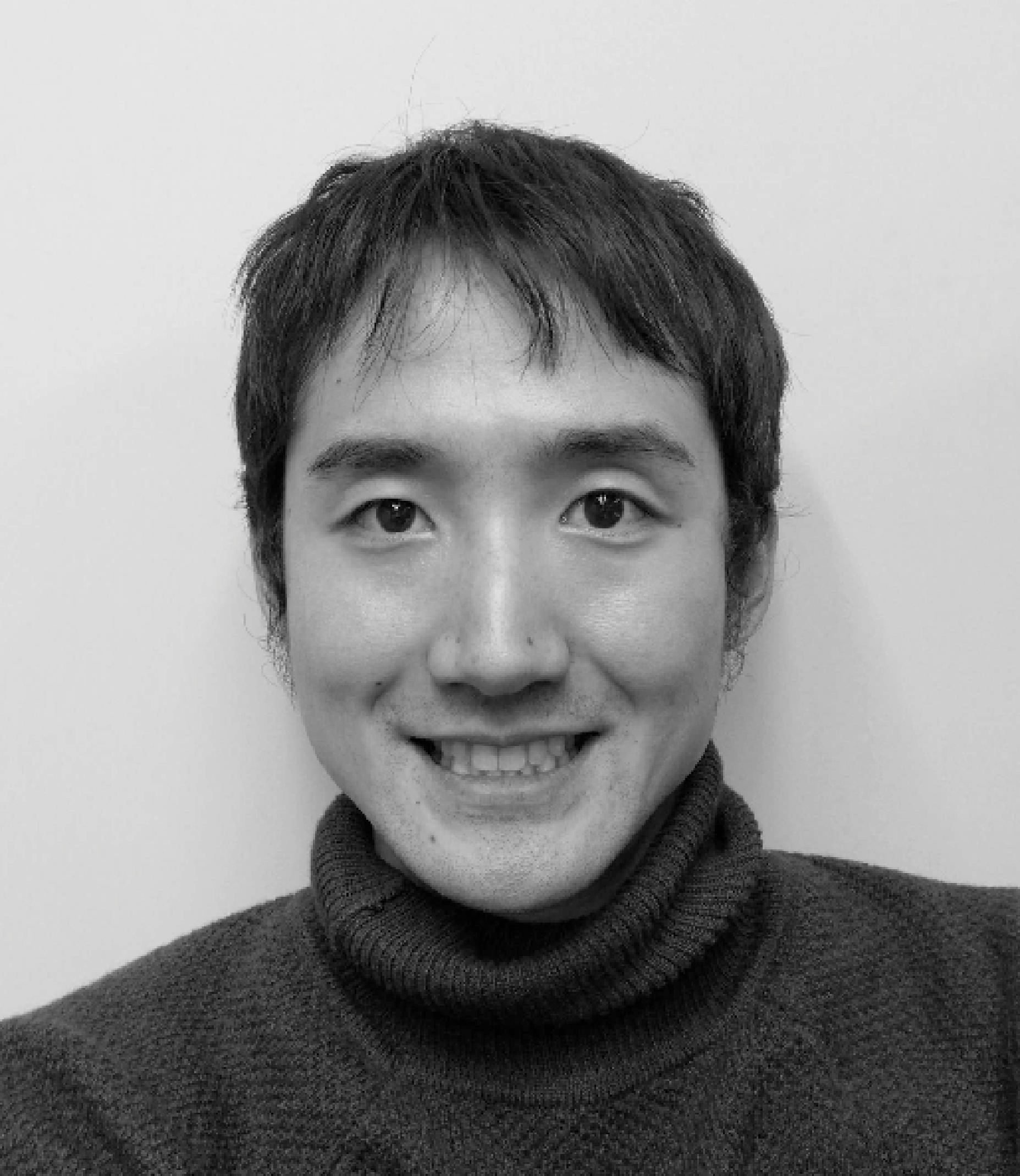}}]{Yusuke Sakemi}
received the M.S. and Ph.D. degrees in physics from The University of Tokyo, Japan in 2012 and 2015, respectively.
He is currently a researcher at NEC Corporation, Japan,  
and a Private Sector Collaborative Researcher at the Institute of Industrial Science, The University of Tokyo, Japan.
His current research interests include neuromorphic engineering and machine learning.
\end{IEEEbiography}

\begin{IEEEbiography}[{\includegraphics[width=1in,height=1.25in,clip,keepaspectratio]{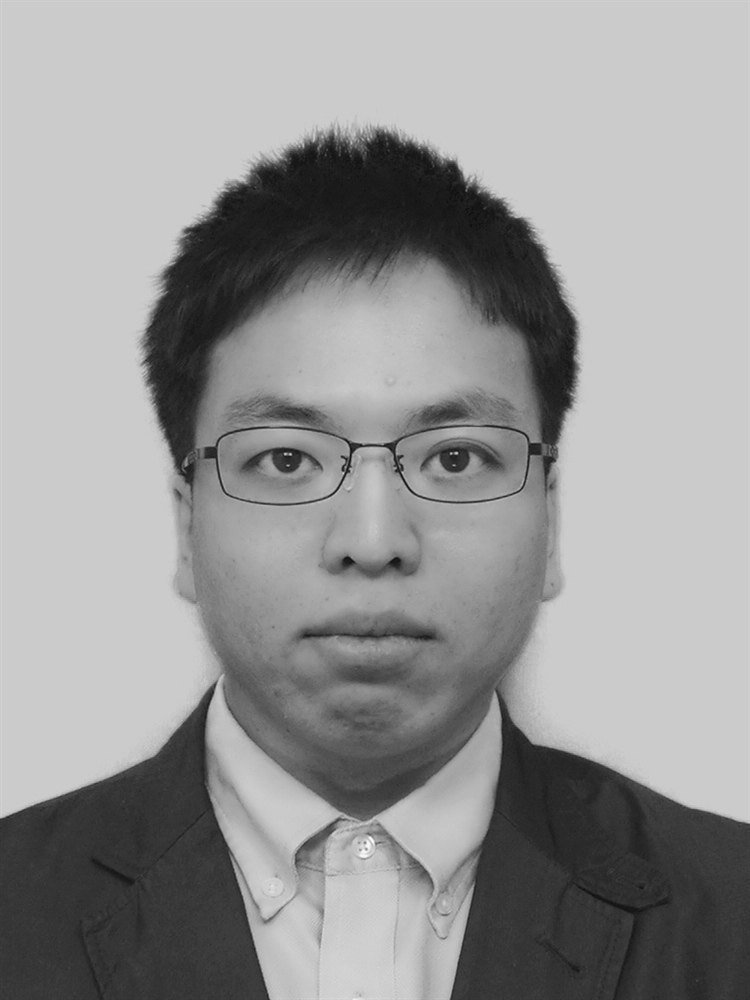}}]{Kai Morino}
received the B.Eng in 2008 and the Master of Informatics in 2010 from Kyoto University, Kyoto, Japan, and the Doctor of Information Science and Technology in 2013 from the University of Tokyo, Tokyo, Japan. He is currently an Associate Professor of Interdisciplinary Graduate School of Engineering Sciences, Kyushu University, Fukuoka, Japan. His research interests include nonlinear dynamical systems and data mining.
\end{IEEEbiography}

\begin{IEEEbiography}[{\includegraphics[width=1in,height=1.25in,clip,keepaspectratio]{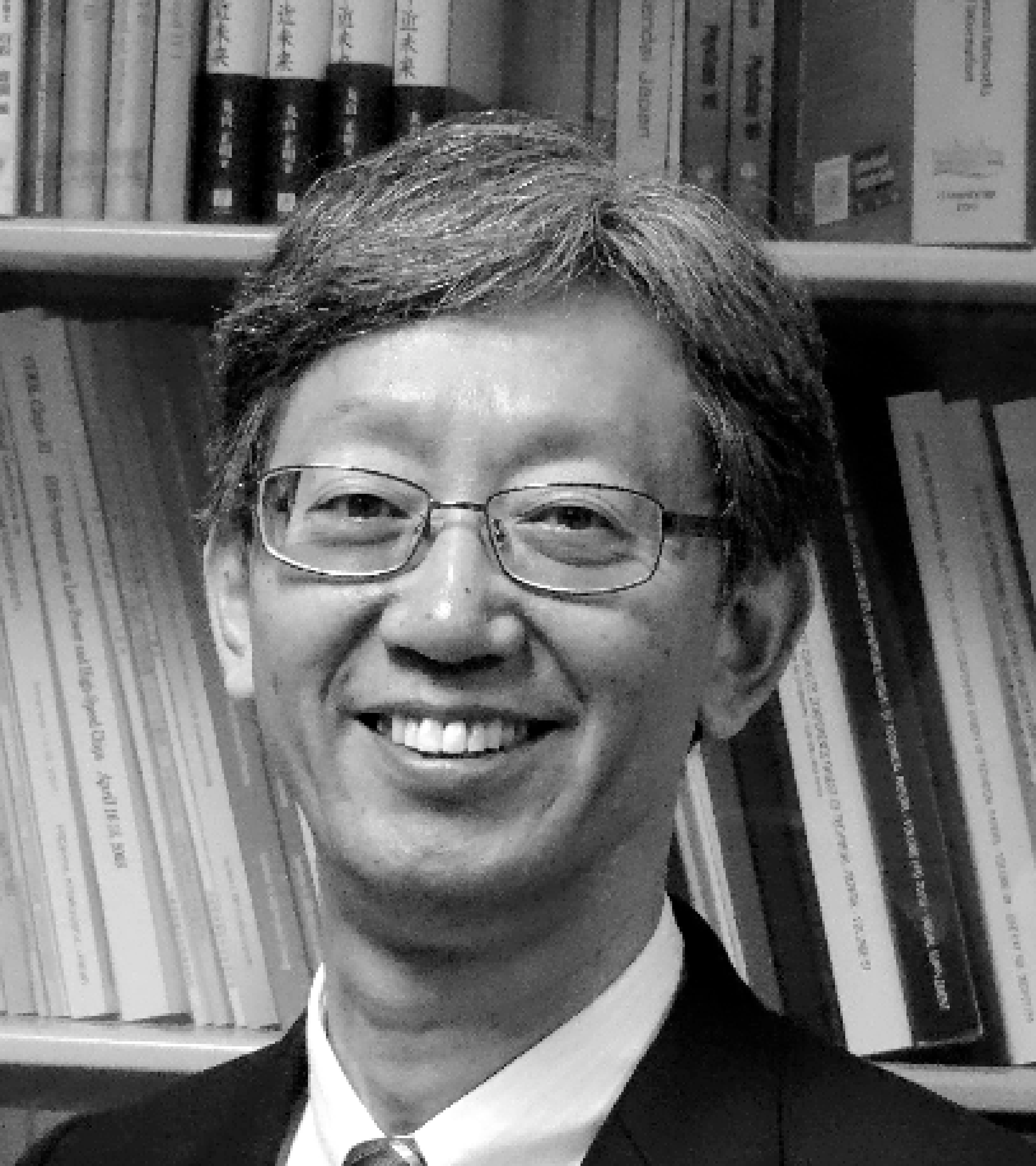}}]{Takashi Morie (M’04)}
received the B.S. and M.S. degrees in Physics from Osaka University, Osaka, Japan, and the Dr. Eng. degree from Hokkaido University, Sapporo, Japan, in 1979, 1981 and 1996, respectively. From 1981 to 1997, he was a member of the Research Staff at Nippon Telegraph and Telephone Corporation (NTT). From 1997 to 2002, he was an associate professor of the department of electrical engineering, Hiroshima University, Higashi-Hiroshima, Japan. Since 2002 he has been a professor of Graduate School of Life Science and Systems Engineering, Kyushu Institute of Technology, Kitakyushu, Japan. His research has been concerned with VLSI implementation of neural networks, brain-like systems and related new functional devices. 
\end{IEEEbiography}

\begin{IEEEbiography}[{\includegraphics[width=1in,height=1.25in,clip,keepaspectratio]{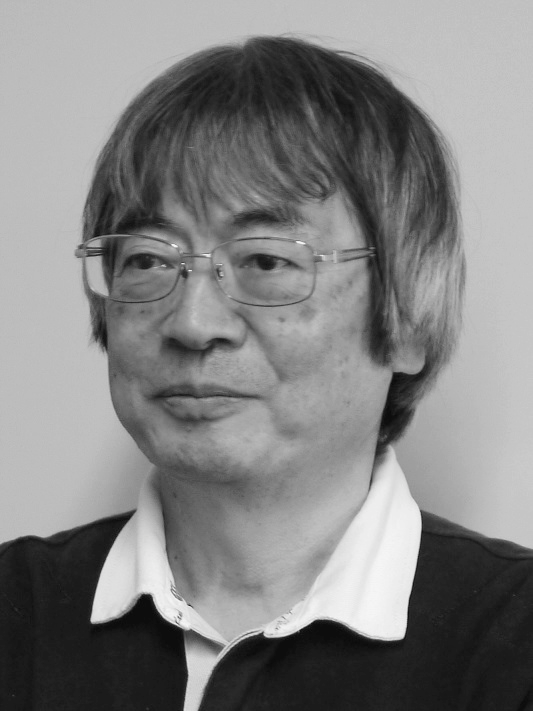}}]{Kazuyuki Aihara}
received the B.E. degree of electrical engineering in 1977 and the Ph.D. degree of electronic engineering 1982 from the University of Tokyo, Japan. Currently, he is Professor of Institute of Industrial Science, Deputy Director of International Research Center for Neurointelligence, Professor of Graduate School of Information Science and Technology, and Professor of Graduate School of Engineering at the University of Tokyo, and Special Advisor of RIKEN Center for Advanced Intelligence Project (AIP). His research interests include mathematical modeling and analysis of such complex systems as the brain and cancer, and nonlinear analysis of complex big data.
\end{IEEEbiography}
\fi







\end{document}